\documentclass[10pt,twocolumn,letterpaper]{article}

\usepackage{cvpr}              

\usepackage[dvipsnames]{xcolor}

\definecolor{cvprblue}{rgb}{0.21,0.49,0.74}
\usepackage[pagebackref,breaklinks,colorlinks,citecolor=cvprblue]{hyperref}
\definecolor{Gray}{gray}{0.7}
\definecolor{DarkGray}{gray}{0.5}
\newcommand{\jyp}[1]{{\color[rgb]{0,0,0}{#1}}}

\usepackage[accsupp]{axessibility} 

\usepackage{hyperref}
\usepackage{algorithm}
\usepackage{algpseudocode}
\usepackage{multirow}
\usepackage{array}
\usepackage{colortbl}
\usepackage{graphicx}
\usepackage{pifont}
\newcommand{\cmark}{\ding{51}}%

\definecolor{tabhighlight}{HTML}{e5e5e5}
\newcommand{\tableCellHeight}{1}
\newcommand{\tabstyle}[1]{
  \setlength{\tabcolsep}{#1}
  \renewcommand{\arraystretch}{\tableCellHeight}
  \centering
  \small
}
\def\paperID{9040} 
\def\confName{CVPR}
\def\confYear{2024}

\title{Prompt Learning via Meta-Regularization}

\author{Jinyoung Park, Juyeon Ko, Hyunwoo J. Kim\thanks{is the corresponding author.}\\
Department of Computer Science and Engineering, Korea University\\
{\tt\small \{lpmn678, juyon98, hyunwoojkim\}@korea.ac.kr}\\
}

\newcommand{\omitme}[1]{}
\newcommand{\hjk}[1]{{\color[rgb]{0,0,0}{#1}}}

\newcommand{\Mc}{\mathcal{M}}
\newcommand{\Lc}{\mathcal{L}}
\newcommand{\Tc}{\mathcal{T}}

\newcommand{\pb}{\mathbf{p}}

\newcommand{\phib}{\boldsymbol{\phi}}

\newcommand{\Dc}{\mathcal{D}}
\newcommand{\Dtr}{D^{\text{tr}}}
\newcommand{\Dval}{D^{\text{val}}}
\newcommand{\Thetab}{\boldsymbol{\Theta}}
\newcommand{\gb}{\boldsymbol{g}}
\newcommand{\gbreg}{\boldsymbol{g}_\text{reg}}

\DeclareMathOperator*{\argmin}{arg\,min}

\def\eg{\emph{e.g}.}
\def\ie{\emph{i.e}.}
\def\xb{\mathbf{x}}
\def\wb{\mathbf{w}}

\def\zb{\mathbf{z}}

\newcommand{\simfunc}{\text{sim}}

\makeatletter
\newcount\my@repeat@count
\newcommand{\myrepeat}[2]{%
  \begingroup
  \my@repeat@count=\z@
  \@whilenum\my@repeat@count<#1\do{#2\advance\my@repeat@count\@ne}%
  \endgroup
}
\makeatother

\begin{document}
\maketitle
\begin{abstract}
Pre-trained vision-language models have shown impressive success on various computer vision tasks with their zero-shot generalizability.
Recently, prompt learning approaches have been explored to efficiently and effectively adapt the vision-language models to a variety of downstream tasks.
However, most existing prompt learning methods suffer from task overfitting since the general knowledge of the pre-trained vision language models is forgotten while the prompts are finetuned on a small data set from a specific target task.  
To address this issue, we propose a \textbf{Pro}mpt \textbf{Meta}-\textbf{R}egularization~(ProMetaR) to improve the generalizability of prompt learning for vision-language models.
Specifically, ProMetaR meta-learns both the regularizer and the soft prompts to harness the task-specific knowledge from the downstream tasks and task-agnostic general knowledge from the vision-language models.
Further, ProMetaR augments the task to generate multiple virtual tasks to alleviate the meta-overfitting.
In addition, we provide the analysis to comprehend how ProMetaR improves the generalizability of prompt tuning in the perspective of the gradient alignment.
Our extensive experiments demonstrate that our ProMetaR improves the generalizability of conventional prompt learning methods under base-to-base/base-to-new and domain generalization settings.
The code of ProMetaR is available at \href{https://github.com/mlvlab/ProMetaR}{https://github.com/mlvlab/ProMetaR}.
\end{abstract}
\vspace{-5mm}
\section{Introduction}
Foundational vision-language models (VLMs) have established their precedence in various computer vision applications such as object detection~\cite{du2022learning,gu2022open,feng2022promptdet,zhong2022regionclip}, image classification~\cite{radford2021learning,singh2022flava,zhai2022lit}, segmentation~\cite{luddecke2022image}, and captioning~\cite{li2023blip,mokady2021clipcap,zhang2023llama}.
Represented by CLIP~\cite{radford2021learning} and ALIGN~\cite{jia2021scaling}, these models are pre-trained on millions of image-text pairs with contrastive loss, creating a shared, well-aligned joint embedding space for vision and language.
They have demonstrated their generalization abilities in zero-shot image recognition and object detection.

\begin{figure}[!t]
\centering
\includegraphics[width=1.0\columnwidth]{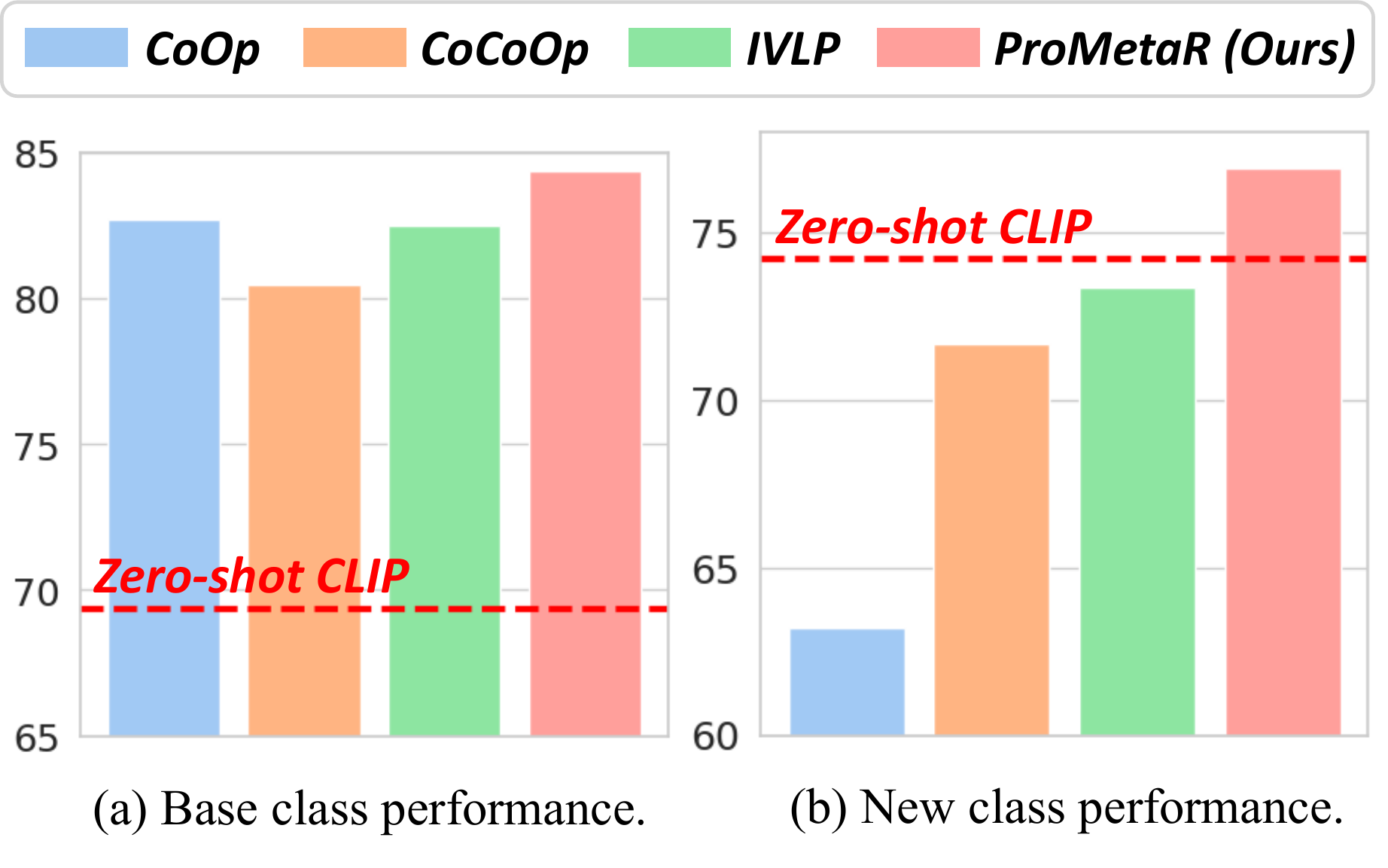}
\caption{Performance comparison of ProMetaR with prompt learning methods~(Zero-shot CLIP, CoOp, CoCoOp, IVLP (base method), and ProMetaR (Ours)) under the base-to-base/base-to-new setting. 
We measure average accuracy on the base classes~(a) and new classes~(b) over 11 datasets.
The red dotted line indicates the performance of the zero-shot CLIP. 
}
\label{fig:polygon}
\end{figure}
Despite the effectiveness of VLMs on zero-shot image recognition, they suffer from time-consuming manual text prompting for each task, which is inefficient and requires human efforts and prior knowledge.
Prompt tuning methods such as Context Optimization~(CoOp)~\cite{zhou2022learning} have arisen as a new paradigm that uses a small number of learnable vectors~(soft prompts) instead of manual prompting.
They efficiently and effectively adapt models to downstream tasks by optimizing only a small number of learnable vectors~(soft prompts) while keeping VLMs frozen.
In recent, some works~\cite{khattak2023maple,lee2023read} further enhance the performance by applying prompt tuning to both image and text modalities.
Prompt tuning methods enhance \textit{traditional} generalization capabilities showing good performance on trained tasks with only a few samples.
However, as the soft prompts tend to prioritize task-specific knowledge, they easily overfit the target task and show poor \textit{task} generalization abilities.
In other words, they have difficulty in generalizing on new tasks, resulting in worse performance than CLIP in data-deficient settings.
From Figure~\ref{fig:polygon}, standard prompt learning methods~(CoOp, CoCoOp, and IVLP) show worse performance than zero-shot CLIP on the unseen~(new) classes during the training, while they perform well on the seen~(base) classes.

One remedy to alleviate the \textit{task} overfitting is learning the learnable prompts with the regularizer.
However, the regularizers are not always beneficial for all the tasks, and it is nontrivial to manually balance the strength of the downstream loss~(\ie, contrastive loss) and regularizer for each task.
So, we propose a framework named ProMetaR~(\textbf{Pro}mpt learning via \textbf{Meta} \textbf{R}egularization) that jointly meta-learns the regularizer and soft prompts to improve the generalizability of the prompt tuning.
Specifically, ProMetaR learns to modulate the gradients of the regularizer to automatically learn effective regularization with a learnable gradient modulation function.
This can be viewed as a bi-level optimization, which can be solved with the meta-learning algorithm.
The representations learned through the meta-learning algorithms are at a high risk of suffering from \emph{meta-overfitting}, meaning that the meta-parameters are overfitted to a small set of validation data (also referred to as meta-data).
To address this issue, we present task augmentation to generate diverse virtual tasks by augmenting the validation set. 
We also show how ProMetaR improves the generalizability of existing prompting methods from the perspective of gradient alignments.

Our extensive experiments validate the effectiveness of ProMetaR under the \textcolor{black}{base-to-base/base-to-new generalization and domain generalization settings} over 11 image recognition datasets and four variants of Imagenet datasets.
In the base-to-base/base-to-new generalization settings~(Figure~\ref{fig:polygon}), our ProMetaR outperforms existing prompt learning methods on 11 image recognition datasets on the both base classes and new classes. 
It also outperforms CLIP on the new classes while improving the performance on the base classes.
These indicate that ProMetaR is effective in both \textit{traditional} generalization and \textit{task} generalization. 
Further, ProMetaR demonstrates its competitive performance under the domain generalization setting.  
We also show that our ProMetaR is applicable to various prompting methods as a general training scheme.

The \textbf{contribution} of our work can be summarized as:
\begin{itemize}
    \item We propose ProMetaR, a prompt learning framework for improving the generalizability of the prompt optimization methods.
    ProMetaR meta-learns both the regularizer and learnable prompts, incorporating task augmentation for more effective meta-learning.
    \item \textcolor{black}{We provide the theoretical analysis of how our ProMetaR improves the generalizability of prompt learning approaches. 
    }
    
    \item Our experiments demonstrate the effectiveness and robustness of ProMetaR under the base-to-base/base-to-new settings and domain generalization. Our ProMetaR significantly improves the base prompting methods on the seen~(base) and unseen~(new) tasks. 
\end{itemize}

\section{Related works}
\vspace{1mm}
\noindent\textbf{Meta-Learning.}
The goal of meta-learning, as known as \textit{learning to learn}, is to efficiently and effectively adapt to new tasks by leveraging past learning experiences~\cite{hospedales2021meta}.
Applications of learning to learn include learning loss functions~\cite{DBLP:conf/nips/ShuXY0ZXM19,DBLP:conf/nips/BalajiSC18,DBLP:conf/icpr/BechtleMCGRSM20}, learning initialization for task adaptation~\cite{finn2017model}, and few-shot learning~\cite{koch2015siamese,sung2018learning,snell2017prototypical}.
Meta-learning algorithms are typically categorized into three types: metric-based methods~\cite{snell2017prototypical,sung2018learning,vinyals2016matching,lee2019meta}, memory-based methods~\cite{mishra2018simple,santoro2016meta,munkhdalai2017meta,munkhdalai2018rapid,hochreiter2001learning}, and gradient-based methods~\cite{ravi2016optimization,grant2018recasting,nichol2018first,li2018learning}.
After Model-agnostic meta-learning (MAML)~\cite{finn2017model} has been proposed, gradient-based approaches have been actively explored.
But, the gradient-based approaches are often prone to meta-overfitting due to insufficient meta-training tasks~\cite{antoniou2018train,yao2021meta,zintgraf2019fast,hwang2020self,hwang2021self,ko2023meltr}.
Inspired by these works, ProMetaR automatically learns the effective regularization in a meta-learning manner for the generalizability of the prompting methods and address the meta-overfitting via task augmentation.

\vspace{1mm}
\noindent\textbf{Regularization.}
Regularization is a conventional technique to prevent neural networks from overfitting and enhance the generalization.
Conventional regularization methods include constraint-based approaches like weight decay~\cite{loshchilov2019decoupled,zhang2019three}, and input-dependent or parameter-dependent approaches such as ensembling~\cite{ilharco2022patching,wortsman2022robust}, dropout~\cite{srivastava2014dropout}, and data augmentation~\cite{zhang2017mixup,yun2019cutmix,verma2018manifold,uddin2021saliencymix,kim2021co,choi2023tokenmixup,park2021metropolis}.
In this work, we present learning to regularize the soft prompts and task augmentation to improve the \textit{traditional} generalization and \textit{task} generalization abilities.  

\vspace{1mm}
\noindent\textbf{Prompt Learning in Vision-Language Models.}
Prompt learning has proven to be an effective technique in various natural language processing tasks~\cite{li2021prefix,lester2021power,liu2023gpt}.
Inspired by this success, prompt learning in vision-language models has also been explored~\cite{zhou2022conditional,lu2022prompt}.
Specifically, CoOp~\cite{zhou2022learning} introduces learnable prompts, or soft prompting, which enables efficient finetuning and adaptation of CLIP~\cite{radford2021learning} text encoder.
VPT~\cite{jia2022visual} proposes to optimize the prompts in the Vision Transformer (ViT)~\cite{dosovitskiy2021image}.
Recently, a line of works~\cite{zang2022unified,khattak2023maple,cho2023distribution} presents multimodal prompt tuning methods by combining the vision and language prompts.
However, many prompt learning approaches in VLMs suffer from the over-fitting issue.
Some works have been proposed to address it.
For example, ProGrad~\cite{zhu2023prompt} regularizes the learning process by aligning the update of the soft prompts to the the task-agnostic general knowledge of the VLMs with the gradient alignment.
UNIGRAM~\cite{li2023gradient} meta-learns the prompt initialization with a large scale of external data to alleviate the generalizability degradation. 
PromptSRC~\cite{khattak2023self} regulates the prompt with mutual agreement maximization and self-ensemble.
Our ProMetaR meta-learns both learnable prompts and regularization to improve the generalizability without using any external data.

\section{Method}
 \begin{figure*}[!ht]
\includegraphics[width=1.0\textwidth]{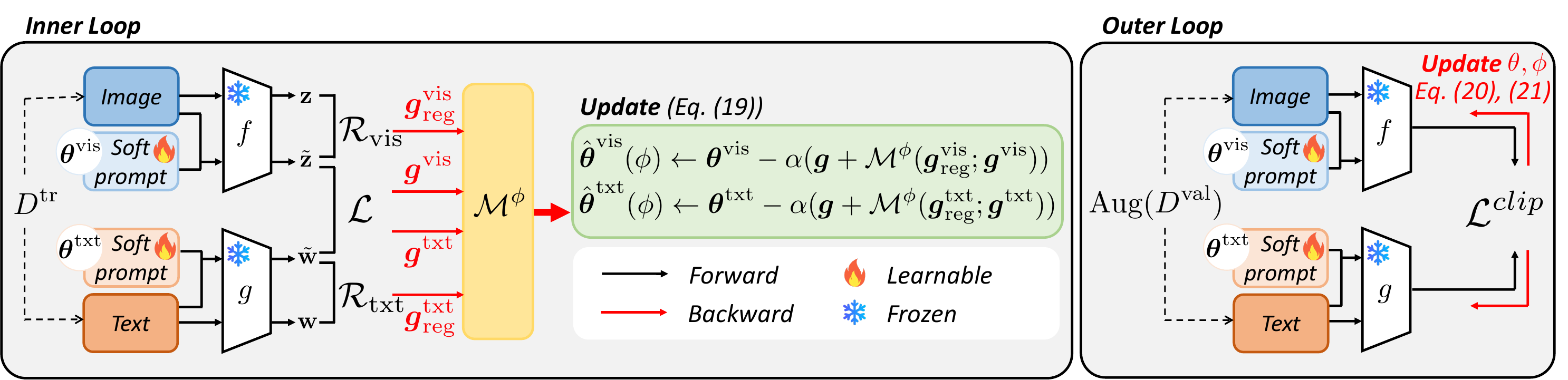}
\caption{
ProMetaR learns the soft prompts~$\Theta = \left\{\boldsymbol{\theta}^{\text{vis}},\boldsymbol{\theta}^{\text{txt}} \right\}$ with meta-regularization to generalize well on the new tasks without losing the generalizability of the pretrained VLMs~(\eg, CLIP).
In the inner-loop~(Eq.~\eqref{eq:inner}), we adapt the soft prompts $\Theta$ with the gradients $\boldsymbol{g}$ of the loss $\mathcal{L}$ and modulated gradients $\boldsymbol{g}_{\text{reg}}=\mathcal{M}^{\boldsymbol{\phi}}\left(\boldsymbol{g}_{\text{reg}};\boldsymbol{g} \right)$.
In the outer-loop~(Eq.~\eqref{eq:outer1}, \eqref{eq:outer2}), the soft prompts $\boldsymbol{\Theta}$ and the gradient modulation function $\boldsymbol{\phi}$ are updated on the augmented validation set $D^{\text{val}}$.
The image encoder $f$ and text encoder $g$ of the pretrained vision-language models are frozen during the training phase. 
%
}
\label{fig:main}
\end{figure*}
\hjk{
We present our ProMetaR (\textbf{P}rompt learning via \textbf{Meta} \textbf{R}egularization) to address the limitations of prompt learning in small data regimes.
Our novel framework automatically learns effective regularization via meta-learning. We will refer to it as Meta Regularization.
Remarkably, the proposed method effectively improves the performance in not only base tasks~(\textit{traditional} generalization) but also new tasks (\emph{task} generalization) to address the \textit{task }overfitting problem.
We first introduce the background of prompt tuning methods for the vision-language models and the meta-learning.
Second, we propose a prompt learning mechanism via meta-regularization to address the over-fitting problems of the prompting approaches.
Finally, we provide the theoretical analysis of our ProMetaR to demonstrate how it enhances the prompt tuning methods.
}
\subsection{Preliminaries}
\hjk{

\noindent\textbf{Prompt tuning for VLMs.}
CLIP~\cite{radford2021learning} provides a well-aligned image-text joint embedding space.
The pre-trained CLIP image encoder $f$ and text encoder $g$ can be used for zero-shot visual recognition by constructing the \emph{hard prompt}.
Specifically, CLIP employs text prompts $\pb_{y}$ generated by hand-crafted templates~(\eg, $``\texttt{A photo of a [CLASS]}"$). 
Then, the prediction probability can be calculated using the visual embeddings $\zb = f(\mathbf{x})$ and textual embeddings $\wb_y = g(\pb_y)$.
Given $N_c$ classes, the predicted probability of image $\mathbf{x}$ to be class $y$ is given as:
\begin{equation}
    p\left(y|\mathbf{x} \right) = \frac{\exp \left(\text{sim}\left(\mathbf{z},\mathbf{w}_y\right)/\tau \right)}{\sum_{j=1}^{N_c}\exp \left(\simfunc\left(\zb, \wb_j\right)/\tau \right)},
\end{equation}
where $\simfunc \left(\cdot, \cdot\right)$ denotes the cosine similarity, $\mathbf{w}_y$ is the textual embedding of the class $y$, and $\tau$ is the temperature.
Even though hard prompts considerably improve CLIP's performance, this technique requires manual efforts to find effective hand-crafted templates for each task, namely, `prompt engineering'. 
Instead of manually optimizing hard prompts, `prompt tuning', also known as `prompt learning', approaches have been proposed to learn context vectors for 
the textual and/or visual prompts, namely \emph{soft prompts}~\cite{zhou2022learning,jia2022visual}.
Concretely, by inserting $N_t$ learnable textual prompts $\boldsymbol{\theta}^{\text{txt}} = \left\{\theta_1^{\text{txt}}, \cdots, \theta_{N_t}^{\text{txt}} \right\}$ and $N_v$ visual prompts $\boldsymbol{\theta}^{\text{vis}} = \left\{\theta_1^{\text{vis}}, \cdots, \theta_{N_v}^{\text{vis}} \right\}$, the textual embedding $\tilde{\wb}_{y}$ for class $y$ and visual embedding $\tilde{\zb}$ are obtained as follows:
\begin{gather}
    \tilde{\wb}_{y} = f\left( \left[\theta_1^{\text{txt}}, \dots, \theta_{N_t}^{\text{txt}}, \boldsymbol{c}_y \right] \right), \\
    \tilde{\zb} = g\left( \left[\mathtt{CLS}, \theta_1^{\text{vis}}, \dots, \theta_{N_v}^{\text{vis}}, \boldsymbol{E} \right] \right),
\end{gather}
where $\boldsymbol{c}_y$ is the word embedding of class $y$, $\mathtt{CLS}$ denotes the class token, and $\boldsymbol{E}$ is the image patch embeddings.
With the weights of the visual encoder $f$ and text encoder $g$ frozen, the prompts are optimized with the contrastive loss:
\begin{equation}
    \mathcal{L} = - \sum_i \mathbf{y}_{i} \log \frac{\exp \left(\simfunc\left(\tilde{\zb}_i,\tilde{\wb}_{i}\right)/\tau \right)}{\sum_{j=1}^{N_c}\exp \left(\simfunc\left(\tilde{\zb}_i, \tilde{\wb}_j\right)/\tau \right)},
\end{equation}
where $\mathbf{y}_{i}$ denotes the one-hot vector for the class of the input $\mathbf{x}_i$.
With the soft prompts, prompt tuning minimizes manual efforts, and it improves CLIP's performance in the downstream tasks.
However, since existing prompt tuning methods tend to focus on \textit{task-specific knowledge}, they often suffer from the overfitting problem, necessitating proper regularization, especially in small data regimes.

\vspace{1mm}
\noindent\textbf{Meta-learning.}
The goal of meta-learning, commonly referred to as `learning-to-learn,' is to design models that can quickly adapt to new tasks with small data
by leveraging past learning experiences across multiple tasks~\cite{hospedales2021meta}.
Let $\Dc$ denote a meta-training set that consists of training and validation sets across tasks $\Tc$, \ie, 
$\Dc = \{ \{D_i^\text{tr}, D_i^\text{val} \} \}_{i \in \Tc}$, where $D_i^\text{tr}$, and $D_i^\text{valid}$ are
the \textit{traditional} training and validation sets of $i$-th task.
Then, meta-learning can be formulated as a bi-level optimization problem given as:
\begin{align}
\label{eq:bilevel-1}
    \min_{\phi}  &\; \sum_{i \in \Tc}  \mathcal{L}_{\text{valid}}({\theta^*_i}(\phi); D^{\text{val}}_i)   \\
\label{eq:bilevel-2}
    \text{s.t. } &\; {\theta_i^*}(\phi)  = \argmin_{\theta_i} \mathcal{L}_{\text{train}} (\theta_i;\phi, D^{\text{tr}}_i), \forall i \in \Tc,
\end{align}
where $\mathcal{L}_{\text{valid}}$ and $\mathcal{L}_{\text{train}}$ denote the losses for the upper- and lower-level optimization problems, and $\phi_i, \theta_i$ are task-specific parameters for $i$-th task and meta-parameters, respectively.
The lower-level optimization in Eq.~\eqref{eq:bilevel-2} does the task-specific adaption/training leveraging learning experiences encoded in 
the meta-parameters $\phi$ and training set $D_i^\text{tr}$. 
The upper-level optimization in Eq.~\eqref{eq:bilevel-1} searches for meta-parameters $\phi$ that improve the overall validation losses of trained task-specific parameters $\theta_i^*(\phi)$.

A seminal work, MAML~\cite{finn2017model}, can be derived from the formulation above. 
MAML aims at learning good initialization that is efficiently adaptable to new tasks.
Let $\phi$ denote the initialization of model parameters.
With task-specific loss function $\Lc_{i}$ and the approximation of lower optimization by one-step update~(Eq.~\eqref{eq:maml-2}), 
the meta-learning formulation in~(Eq.~\eqref{eq:bilevel-1}) is converted into MAML's formulation given as:
\begin{align}
\label{eq:maml-1}
\min_\phi & \sum_i \mathcal{L}_{i}({\hat{\theta}}_i(\phi); D_i^{\text{val}}) \\
\label{eq:maml-2}
\text{s.t. }& \hat{{\theta}}_i(\phi) = \phi - \alpha \nabla_\phi \mathcal{L}_{i}(\phi; {D}_i^\text{tr}), \forall i \in  \Tc,
\end{align}
where $\alpha$ denotes the step size to adapt the initialization $\phi$ to $i$-th task.
The approximation by one-step update in Eq.~\eqref{eq:maml-2} enables efficient optimization without the necessity of iterative optimization for lower optimization. 
}

\subsection{Prompt learning via meta-regularization}
\hjk{
We propose a novel framework named Prompt Learning via Meta-Regularization~(ProMetaR) to improve the generalizability of existing prompt tuning methods.
Our framework adopts meta-learning approaches to learn both soft prompts and regularizers. 
In addition, we incorporate \textit{task augmentation} into our framework to generate diverse tasks to alleviate the \textit{meta-overfitting}.
Figure~\ref{fig:main} delineates the overall meta-learning pipeline of the proposed method.

Prompt tuning optimizes prompts to adapt pre-trained models, \eg, a Vision-Language Models~(VLMs), to the specific tasks by minimizing a loss:
\begin{equation}
    \min_{\boldsymbol{\Theta}} \mathcal{L}\left(\boldsymbol{\Theta};D^{\text{tr}} \right),
\end{equation}
where $\boldsymbol{\Theta}=\{\boldsymbol{\theta}^{\text{txt}}, \boldsymbol{\theta}^{\text{vis}} \}$ denotes the learnable prompts and $D^{\text{tr}}$ is the training set of the target downstream task.
Since the goal of prompt tuning is the sample-efficient adaptation of the pre-trained models, 
the training set for prompt tuning is usually small. 
Thus, prompt tuning methods often suffer from overfitting, showing inferior performance compared to even zero-shot VLMs.
To address this problem, we introduce a regularizer $\mathcal{R}$ that penalizes
large changes in representations as
\begin{equation}
    \mathcal{R}_{\text{vis}} = \sum_i\left\lvert \tilde{\mathbf{z}}_i-\mathbf{z}_i \right\rvert, \quad \mathcal{R}_{\text{txt}} = \sum_j \left\lvert \tilde{\mathbf{w}}_j-\mathbf{w}_j \right\rvert,
\end{equation}
where $\mathbf{z}, \mathbf{w}$ denote original visual and textual embeddings, while $\tilde{\mathbf{z}}, \tilde{\mathbf{w}}$ represent corresponding embeddings obtained with prompts $\Theta$.
Then, we have:
\begin{equation}
\label{eq:loss+reg}
    \min_{\boldsymbol{\Theta}} \mathcal{L}\left(\boldsymbol{\Theta};D^{\text{tr}} \right) + \lambda \mathcal{R}(\boldsymbol{\Theta}; D^{\text{tr}}), 
\end{equation}
where $\lambda \in \mathbb{R}_{+}$ is the regularization strength and $\mathcal{R}$ unifies $\mathcal{R}_{\text{vis}}$, and $\mathcal{R}_{\text{txt}}$.

However, the regularizer may not always be helpful and manually adjusting the strength of the regularizer, is nontrivial.
So, we learn the regularizer to automatically balance it with the main loss, which can be formulated as a bi-level optimization given as:
\begin{align}
    \label{eq:remark_eq}
        \min_{\Thetab,\boldsymbol{\phi}} & \; \mathcal{L} \left({\Thetab^*} \left(\boldsymbol{\phi} \right); {D}^{\text{val}} \right) \\
        \text{s.t. } &  {\boldsymbol{\Theta}^*}\left(\boldsymbol{\phi} \right) = \argmin_{\boldsymbol{\Theta}}\mathcal{L}\left( \boldsymbol{\Theta} ;{D}^{\text{tr}}\right)+ \mathcal{R}^{\boldsymbol{\phi}}\left( \boldsymbol{\Theta} ;{D}^{\text{tr}}\right), \nonumber
\end{align}
where $\Thetab$ is a meta-parameter for the better adaptation, and $\boldsymbol{\phi}$ is a also meta-parameter to learn the strength of regularizer $\mathcal{R}$.
Similar to Eq.~\eqref{eq:maml-2} in MAML, using the one-step update approximation, 
Eq.~\eqref{eq:remark_eq} can be rewritten as 
\begin{align}
\label{eq:meta_kd}
&\min_{\boldsymbol{\Theta}, \boldsymbol{\phi}} \mathcal{L}\left(\hat{\boldsymbol{\Theta}}\left(\boldsymbol{\phi} \right); D^{\text{val}}  \right)\\
\label{eq:meta_kd-2}
&\text{s.t. } \hat{\boldsymbol{\Theta}}\left(\boldsymbol{\phi} \right) = \boldsymbol{\Theta} - \alpha \left(\gb + \Mc^{\phib} \left( \gb_{\text{reg}} ;\gb  \right)\right),
\end{align}
where $\boldsymbol{g} = \nabla_{\boldsymbol{\Theta}}\mathcal{L}\left( \boldsymbol{\Theta}; \Dtr\right)$ and $\boldsymbol{g}_{\text{reg}} = \nabla_{\boldsymbol{\Theta}}\mathcal{R}\left( \boldsymbol{\Theta}; {D}^{\text{tr}}\right)$, 
and $\mathcal{M}^{\boldsymbol{\phi}}$ is the gradient modulation function with the parameter $\phib$ that 
adaptively adjusts $\gbreg$ considering $\gb$ as:
\begin{equation}
    \mathcal{M}^{\boldsymbol{\phi}}\left(\gbreg ; \gb \right) = \sigma\left(\boldsymbol{m}^{\boldsymbol{\phi}}\right)\odot \boldsymbol{g}_{\text{reg}}, 
\end{equation}
where $\sigma$ is the sigmoid function and $\odot$ is Hadamard product.
The modulation vectors $\boldsymbol{m}^{\boldsymbol{\phi}}$ is computed by $\text{MLP}_\phi\left(\left[ \boldsymbol{g}|| \boldsymbol{g}_{\text{reg}}\right]\right)$ considering the gradients of both the loss and the regularizer.

By learning the regularizer, we have addressed the overfitting problem of the prompt learning methods.
We further extend our framework to boost generalization performance in new tasks (\textit{task} generalization) by generating diverse tasks. 
To this extent, we incorporate \textit{task augmentation} into our framework as:
\begin{equation}
\label{eq:meta_aug}
\begin{split}
        \min_{\boldsymbol{\Theta}, \boldsymbol{\phi}} & \; \mathbb{E} \; \mathcal{L}\left(\hat{\boldsymbol{\Theta}}\left(\boldsymbol{\phi} \right); \text{Aug}\left(\Dval \right)  \right)\\
        \text{s.t. }& \; \hat{\boldsymbol{\Theta}}\left(\boldsymbol{\phi} \right) = \boldsymbol{\Theta} - \alpha \left( \boldsymbol{g} + \mathcal{M}^{\boldsymbol{\phi}}\left(\gbreg ; \gb  \right)\right),
\end{split}
\end{equation}
where $\text{Aug}\left(\cdot\right)$ is the task augmentation operation.
The task augmentation generates new labels to augment many tasks, which encourages the parameters to be optimized for diverse tasks.
The augmented task can be viewed as a virtually large meta-validation set with many tasks.
This helps the model generalize on new tasks.

Mixup-based augmentation is one of the augmentation operations that generate new interpolated labels.
In our experiments, task augmentation randomly draws samples from train and validation sets and employs manifold mix~\cite{verma2018manifold} for augmentation.
Specifically, 
given a pair of random samples $\xb_i \in \Dval$ and $\xb_j \in \Dtr$ from validation and training sets, 
we interpolate the last layer features~($\boldsymbol{h}^{(i)}_{\text{val}}, \boldsymbol{h}^{(j)}_{\text{tr}}$) and their labels~($y^{(i)}_{\text{val}}, y^{(j)}_{\text{tr}}$) as:
\begin{equation}
    \hat{\boldsymbol{h}}^{(i)}_{\text{val}} = \rho \boldsymbol{h}^{(i)}_{\text{val}}  + (1-\rho) \boldsymbol{h}_{\text{tr}}^{(j)},
\end{equation}
\begin{equation}
    \hat{y}^{(i)}_{\text{val}} = \rho y^{(i)}_{\text{val}}  + (1-\rho) y^{(j)}_{\text{tr}},
\end{equation}
where $\rho \in \left[0,1 \right]$ is a mixture ratio, which is sampled from the Beta distribution $\text{Beta} \left(\mu, \nu \right)$.

\vspace{1mm}
\noindent\textit{Remarks.} Note that similar to overfitting, meta-learning algorithms often suffer from \textit{meta-overfitting}, especially when the size of the meta-validation set $\{ \Dval_i \}_{i \in \Tc}$ in Eq.~\eqref{eq:bilevel-1} is small~\cite{antoniou2018train,yao2021meta,zintgraf2019fast}.
The size is related to the quantity of both samples and tasks, and their diversity. 
Unfortunately, in prompt tuning benchmarks such as base-to-base/base-to-new generalization and domain generalization settings, only one task is available for training with a small number of samples. 
This setting is challenging and can be seen as `single-task meta-learning'.
In our framework, task augmentation effectively addresses the scarcity of tasks/samples and boosts generalization performance in both a base (seen) task
and a new task.

\vspace{1mm}
\textbf{Overall procedure of ProMetaR.}
Motivated by the episodic training scheme~\cite{vinyals2016matching}, we divide the batch into the training and validation set based on the class of the sample.
To maintain the in-domain performance, we first update the parameters with the conventional gradient descent.  
Then, we update the parameters with the meta-learning. 
 The learnable prompts $\boldsymbol{\Theta}$ are adapted with the gradients of the loss and modulated gradients~(inner-loop): 
\begin{equation}
\label{eq:inner}
    \hat{\Thetab}(\phib) \gets \boldsymbol{\Theta} - \alpha \left(\boldsymbol{g} + \Mc^{\phib}\left(\gbreg  ; \gb \right)\right)
\end{equation}
After the update, the learnable prompts $\boldsymbol{\Theta}$ and gradient modulation function $\boldsymbol{\phi}$ are optimized for performing well on the augmented set~(outer-loop):
\begin{equation}
\label{eq:outer1}
    \boldsymbol{\Theta} \gets \boldsymbol{\Theta} - \beta \nabla_{\boldsymbol{\Theta}}  \mathcal{L}\left(\hat{\Thetab}(\phib); \text{Aug}\left( \Dval\right)  \right),
\end{equation}
\begin{equation}
\label{eq:outer2}
    \phib \gets \phib - \beta  \nabla_{\phib} \mathcal{L}\left(\hat{\Thetab}(\phib);\text{Aug}\left(\Dval \right)  \right),
\end{equation}
where $\alpha, \beta$ are hyperparameters, respectively.
}
\subsection{Analysis of ProMetaR}
\hjk{
We provide the analysis to elucidate how our proposed ProMetaR enhances the generalizability of prompt learning from the standpoint of gradient alignment~\cite{vinyals2016matching}. 
The objective of ProMetaR is to find the optimal soft prompts as follows:
\begin{equation}
\label{eq:metaopt_eq}
    \min_{\boldsymbol{\boldsymbol{\Theta}},\boldsymbol{\phi}} \mathcal{L}\left(\boldsymbol{\boldsymbol{\Theta}}-\alpha \left(\boldsymbol{g}+\mathcal{M}^{\boldsymbol{\phi}}\left(\boldsymbol{g}_{\text{reg}} \right)\right); {D}^{\text{val}} \right),
\end{equation}
where $\boldsymbol{g} = \nabla_{\boldsymbol{\boldsymbol{\Theta}}} \mathcal{L}\left(\boldsymbol{\Theta};D^{\text{tr}}\right), \boldsymbol{g}_{\text{reg}} = \nabla_{\boldsymbol{\boldsymbol{\Theta}}} \mathcal{R}\left(\boldsymbol{\Theta};D^{\text{tr}} \right)$ are the gradients of loss~$\mathcal{L}$ and regularizer $\mathcal{R}$, respectively. 

We can approximate $\mathcal{L}\left(\boldsymbol{x}\right)$ with first-order Taylor expansion.
Given loss $\Lc(\boldsymbol{x})$, its first-order approximation via Taylor expansion is as follows:
\begin{equation}
\label{eq:metaopt}
    \mathcal{L}(\boldsymbol{x}) \approx
    \mathcal{L}\left({\boldsymbol{x}_0} \right) + \nabla_{\boldsymbol{x}}\mathcal{L}\left({\boldsymbol{x}_0} \right)^{\top} \left(\boldsymbol{x} - {\boldsymbol{x}_0} \right),
\end{equation}
where $\boldsymbol{x}_0$ is an arbitrary point and $\boldsymbol{x}$ is a point close to $\boldsymbol{x}_0$.
Assume that we have $\boldsymbol{x} = \boldsymbol{\boldsymbol{\Theta}}-\alpha \left(\boldsymbol{g} + \mathcal{M}^{\phi}\left(\boldsymbol{g}_{\text{reg}} \right)\right)$ and $\boldsymbol{x}_0 = \boldsymbol{\boldsymbol{\Theta}}$.
Then, our objective~(Eq.~\eqref{eq:metaopt_eq}) can be written as:
\begin{equation}
\label{eq:metaopt2}
\begin{split}
     \min_{\boldsymbol{\boldsymbol{\Theta}}, \boldsymbol{\phi}}\, &\mathcal{L}\left(\boldsymbol{\boldsymbol{\Theta}}; {D}^{\text{val}} \right) +\\
    &\nabla_{\boldsymbol{\Theta}}\mathcal{L}\left(\boldsymbol{\boldsymbol{\Theta}} \right)^{\top} \left(-\alpha \left(\boldsymbol{g} + \mathcal{M}^{\phi}\left(\boldsymbol{g}_{\text{reg}} \right) \right)\right).
\end{split}
\end{equation}
Since $\Mc^{\phi}\left(\boldsymbol{g}_{\text{reg}} \right) = \sigma\left(\boldsymbol{m}^{\boldsymbol{\phi}} \right)\odot \boldsymbol{g}_{\text{reg}}$, we can rewrite Eq.~\eqref{eq:metaopt2} as below:
\begin{equation}
\begin{split}
\label{eq:metaopt3}
\min_{\boldsymbol{\boldsymbol{\Theta}},\boldsymbol{\phi}}\,\mathcal{L}\left(\boldsymbol{\boldsymbol{\Theta}}; D^{\text{val}} \right) &-\alpha \biggl(\nabla_{\boldsymbol{\Theta}}\mathcal{L}\left(\boldsymbol{\Theta} \right)^{\top} \boldsymbol{g} \biggr)\\
      & - \alpha \biggl ( \nabla_{\boldsymbol{\Theta}}\mathcal{L}\left(\boldsymbol{\Theta} \right)^{\top} \left(\sigma\left(\boldsymbol{m}^{\boldsymbol{\phi}}\right) \odot \boldsymbol{g}_{\text{reg}}  \right)\biggr).
\end{split}
\end{equation}
This equation has three terms. 
The optimization above implies minimizing (i) the loss on the validation set,  
(ii) maximizing the inner product between the gradients of the losses on the validation set and the training set, and
(iii) maximizing the inner product between the gradient of the validation loss and the regularizer on the training set.
So, these indicate that this optimization prefers a solution/direction where the training and validation gradients agree, which leads to 
better generalization on new tasks. 
In addition, the third term in Eq.~\eqref{eq:metaopt3} plays a role in avoiding the conflict of the update between the task-specific knowledge by tuned prompts and task-agnostic general knowledge provided by original prompts.
From the perspective of the gradient alignment~\cite{zhu2023prompt}, the third term leads to a reduction in the generalization error by aligning the gradients induced by tuned prompts and general knowledge from the original prompts.
So, our proposed ProMetaR enhances the \textit{task} generalization ability as well as \textit{traditional} generalization capability.
}
\section{Experiments}
In this section, we demonstrate the effectiveness of our proposed ProMetaR.
We first introduce datasets, baselines, and implementation details.
Next, we provide the ablation studies to explore the contribution of each component in ProMetaR.
\textcolor{black}{Then, we compare the proposed method with other prompting-based methods 
to evaluate the ability of traditional generalization on seen categories (base-to-base), and task generalization 
to unseen categories (base-to-new) and new datasets (domain generalization).
We also design a task overfitting score and provide analysis to show the efficacy of the proposed method.
}
\subsection{Experimental settings}
\hjk{
We evaluate ProMetaR on base-to-base/base-to-new generalization and domain generalization following other prompting works~\cite{khattak2023maple}.

\vspace{1mm}
\noindent\textbf{Base-to-base/Base-to-new generalization.}
We train the prompts only on the base classes in a 16-shot~(16 images per class) setting and measure the performance of the prompting methods on base and new classes.
In this setting, the model cannot see new classes in the training phase.

\vspace{1mm}
\noindent\textbf{Domain generalization.}
We also validate the effectiveness of our model in a 16-shot on out-of-distribution datasets.
We train the model only using ImageNet dataset~(source) and perform inference on four other variants~(target) of ImageNet dataset.
In other words, the model cannot see target domains in the training phase.

\vspace{1mm}
\noindent\textbf{Datasets.}
For base-to-base/base-to-new class generalization, we evaluate our method on 11 image recognition datasets: ImageNet~\cite{deng2009imagenet}, Caltech101~\cite{fei2004learning}, OxfordPets~\cite{parkhi2012cats}, StanfordCars~\cite{krause20133d}, Flowers102~\cite{nilsback2008automated}, Food101~\cite{bossard2014food}, FGVCAircraft~\cite{maji2013fine}, SUN397~\cite{xiao2010sun}, UCF101~\cite{soomro2012ucf101}, DTD~\cite{cimpoi2014describing}, and EuroSAT~\cite{helber2019eurosat}, following other prompting methods~\cite{khattak2023maple,zhou2022conditional}.
We also evaluate our method on domain generalization settings by setting ImageNet~\cite{deng2009imagenet} as the source dataset. 
The target datasets contain four ImageNet variants: ImageNetV2~\cite{recht2019imagenet}, ImageNet-Sketch~\cite{wang2019learning}, ImageNet-A~\cite{hendrycks2021natural}, and ImageNet-R~\cite{hendrycks2021many}.

\vspace{1mm}
\noindent\textbf{Baselines.}
To validate the effectiveness of our ProMetaR, we use the following baselines: (1) zero-shot CLIP~\cite{radford2021learning}, (2) textual prompt learning approaches: CoOp~\cite{zhou2022learning} and CoCoOp~\cite{zhou2022conditional}, (3) multimodal prompt learning approaches: MAPLE~\cite{khattak2023maple} and RPO~\cite{lee2023read}, (4) prompt learning with regularization and ensemble methods: PromptSRC~\cite{khattak2023self}, (5) prompt learning with the meta-learning: UNIGRAM~\cite{li2023gradient}, \jyp{and (6) our base prompting method: IVLP.}

\vspace{1mm}
\noindent\textbf{Experimental details.}
Following other prompt learning works~\cite{zhou2022conditional,khattak2023self,khattak2023maple}, we use CLIP-ViT-B/16 as the pretrained backbone model and four soft prompting tokens for each modality.
For the base prompt learning method, we use Independent Vision-Language Prompting as a base prompt learning method that optimizes hierarchical prompts on both image and text modalities~\cite{khattak2023maple}. 
In all experiments, we evaluate the performance of the methods in three independent runs (seed 1, 2, and 3) and report average performance following other prompt learning works~\cite{zhou2022conditional,khattak2023self,khattak2023maple}.
}
\subsection{Effectiveness of ProMetaR}
\begin{table}[!t]
     \centering
 \setlength{\tabcolsep}{3pt}
    \resizebox{\columnwidth}{!}{
    \begin{tabular}{c|c c c| ccc }
    \toprule
    & MetaLearn & TaskAug & MetaReg & Base & New & H\\
    \midrule
     (a) & & & & 82.51 & 73.36 & 77.66 \\
     (b) &\cmark & & & 83.51 &73.15 & 77.99\\
     (c) &\cmark & \cmark & & 84.04 & 75.37 & 79.47 \\
     (d) &\cmark & & \cmark & 84.27 & 75.06 & 79.40 \\
     \rowcolor{tabhighlight}
     (e) & \cmark & \cmark & \cmark & \textbf{84.39} & \textbf{76.93} & \textbf{80.49} \\
    \bottomrule
    \end{tabular}
    }
    \caption{Contribution of each component of our ProMetaR. Results are averaged over 11 datasets. H refers to harmonic mean. MetaLearn: meta-learning, TaskAug: Task augmentation to alleviate the meta-overfitting, MetaReg: meta-regularization to learn the regularizer.
    }
    \label{tab:ablations_on_components}
\end{table}

\begin{table*}[t]
  \centering
  \small
  \renewcommand{\arraystretch}{0.9}
    \setlength{\tabcolsep}{1.5pt}
    \resizebox{\textwidth}{!}{
    \begin{tabular}{ >{\arraybackslash}m{12mm}  >{\centering\arraybackslash}m{10mm} | >{\centering\arraybackslash}m{14mm}  >{\centering\arraybackslash}m{14mm}  >{\centering\arraybackslash}m{15mm}  >{\centering\arraybackslash}m{15mm}  >{\centering\arraybackslash}m{14mm}  >{\centering\arraybackslash}m{15mm}  >
    {\centering\arraybackslash}m{15mm} | >
    {\centering\arraybackslash}m{15mm}  >{\centering\arraybackslash}m{15mm}  > {\centering\arraybackslash}m{8mm} }
        \toprule
        \multirow{2}{*}{Dataset} & & CLIP & CoOp  & CoCoOp & MaPLe & RPO & PromptSRC & UNIGRAM & IVLP & \textbf{ProMetaR} & Gain\\
         & & \cite{radford2021learning} & \cite{zhou2022learning}  & \cite{zhou2022conditional} & \cite{khattak2023maple} & \cite{lee2023read} & \cite{khattak2023self} & \cite{li2023gradient} & (Base) & \textbf{(Ours)} & $\Delta$\\
        \midrule
        \midrule
         \rowcolor{tabhighlight} \multicolumn{2}{c|}{\textbf{Avg. Rank}} & 8.18 & 8.55 & 6.73 & 3.64 & 4.55 & 2.73 & 3.82 & 5.27 & \textbf{1.36} & - \\
         \midrule
        \multirow{3}{*}{\shortstack[l]{Average on \\11 datasets}} & Base & 69.34 & 82.69 & 80.47 & 82.28 & 81.13 & 84.26 & 80.34 & 82.51 & \textbf{84.39} & \textcolor{NavyBlue}{+1.88} \\
         & New & 74.22 & 63.22 & 71.69 & 75.14 & 75.00 & 76.10 & 75.92 & 73.35 & \textbf{76.93} & \textcolor{NavyBlue}{+3.58}\\
         & H & 71.70 & 71.66 & 75.83 & 78.55 & 77.78 & 79.97 & 78.07 & 77.66 & \textbf{80.49} & \textcolor{NavyBlue}{+2.83}\\
         \midrule
         \multirow{3}{*}{ImageNet} & Base & 72.43 & 76.47 & 75.98 & 76.66 & 76.60 & 77.60 & 76.60 & 77.39 & \textbf{77.76} & \textcolor{NavyBlue}{+0.37} \\
         & New & 68.14 & 67.88 & 70.43 & 70.54 & \textbf{71.57} & 70.73 & 70.69 & 70.04 & 70.75 & \textcolor{NavyBlue}{+0.71} \\
         & H & 70.22 & 71.92 & 73.10 & 73.47 & 74.00 & 74.01 & 73.53 & 73.53 & \textbf{74.09} & \textcolor{NavyBlue}{+0.56} \\
         \midrule
         \multirow{3}{*}{\shortstack[l]{Caltech\\101}} & Base & 96.84 & 98.00 & 97.96 & 97.74 & 97.97 & 98.10 & 98.07 & \textbf{98.28} & 98.11 & \textcolor{Bittersweet}{-0.17} \\
         & New & 94.00 & 89.81 & 93.81 & 94.36 & 94.37 & 94.03 & \textbf{95.11} & 93.65 & 94.29 & \textcolor{NavyBlue}{+0.64} \\
         & H & 95.40 & 93.73 & 95.84 & 96.02 & 96.03 & 96.02 & \textbf{96.57} & 95.91 & 96.16 & \textcolor{NavyBlue}{+0.25} \\
         \midrule
         \multirow{3}{*}{\shortstack[l]{Oxford\\Pets}} & Base & 91.17 & 93.67 & 95.20 & 95.43 & 94.63 & 95.33 & 94.94 & 95.41 & \textbf{95.57} & \textcolor{NavyBlue}{+0.16} \\
         & New & 97.26 & 95.29 & 97.69 & 97.76 & 97.50 & 97.30 & \textbf{97.94} & 96.31 & 97.43 & \textcolor{NavyBlue}{+1.12} \\
         & H & 94.12 & 94.47 & 96.43 & \textbf{96.58} & 96.05 & 96.30 & 96.42 & 95.86 & 96.49 & \textcolor{NavyBlue}{+0.63} \\
         \midrule
         \multirow{3}{*}{\shortstack[l]{Stanford\\Cars}} & Base & 63.37 & 78.12 & 70.49 & 72.94 & 73.87 & 78.27 & 73.50 & 72.39 & \textbf{78.32} & \textcolor{NavyBlue}{+5.93} \\
         & New & 74.89 & 60.40 & 73.59 & 74.00 & \textbf{75.53} & 74.97 & 75.38 & 73.31 & 75.18 & \textcolor{NavyBlue}{+1.87} \\
         & H & 68.65 & 68.13 & 72.01 & 73.47 & 74.69 & 76.58 & 74.43 & 72.85 & \textbf{76.72} & \textcolor{NavyBlue}{+3.87} \\
         \midrule
         \multirow{3}{*}{\shortstack[l]{Flowers\\102}} & Base & 72.08 & 97.60 & 94.87 & 95.92 & 94.13 & 98.07 & 95.20 & 96.17 & \textbf{98.13} & \textcolor{NavyBlue}{+1.96} \\
         & New & \textbf{77.80} & 59.67 & 71.75 & 72.46 & 76.67 & 76.50 & 76.21 & 73.64 & 77.66 & \textcolor{NavyBlue}{+4.02} \\
         & H & 74.83 & 74.06 & 81.71 & 82.56 & 84.50 & 85.95 & 84.65 & 83.41 & \textbf{86.70} & \textcolor{NavyBlue}{+3.29}\\
         \midrule
         \multirow{3}{*}{Food101} & Base & 90.10 & 88.33 & 90.70 & 90.71 & 90.33 & 90.67 & \textbf{90.84} & 90.53 & 90.80 & \textcolor{NavyBlue}{+0.27} \\
         & New & 91.22 & 82.26 & 91.29 & 92.05 & 90.83 & 91.53 & \textbf{92.12} & 91.66 & 91.89 & \textcolor{NavyBlue}{+0.23} \\
         & H & 90.66 & 85.19 & 90.99 & 91.38 & 90.58 & 91.10 & \textbf{91.48} & 91.09 & 91.34 & \textcolor{NavyBlue}{+0.25}\\
         \midrule
         \multirow{3}{*}{\shortstack[l]{FGVC\\Aircraft}} & Base & 27.19 & 40.44 & 33.41 & 37.44 & 37.33 &  \textbf{42.73} & 32.25 & 37.24 & 42.02 & \textcolor{NavyBlue}{+4.78}\\
         & New & 36.29 & 22.30 & 23.71 & 35.61 & 34.20 & 37.87 & 38.00 & 34.47 & \textbf{38.63} & \textcolor{NavyBlue}{+4.16} \\
         & H & 31.09 & 28.75 & 27.74 & 36.50 & 35.70 & 40.15 & 34.89 & 35.80 & \textbf{40.25} & \textcolor{NavyBlue}{+4.45} \\
         \midrule
         \multirow{3}{*}{SUN397} & Base & 69.36 & 80.60 & 79.74 & 80.82 & 80.60 & 82.67 & 80.43 & 82.63 & \textbf{82.70} & \textcolor{NavyBlue}{+0.07} \\
         & New & 75.35 & 65.89 & 76.86 & 78.70 & 77.80 & 78.57 & 77.91 & 78.40 & \textbf{79.02} & \textcolor{NavyBlue}{+0.62} \\
         & H & 72.23 & 72.51 & 78.27 & 79.75 & 79.18 & 80.52 & 79.15 & 80.46 & \textbf{80.82} & \textcolor{NavyBlue}{+0.36} \\
         \midrule
         \multirow{3}{*}{DTD} & Base & 53.24 & 79.44 & 77.01 & 80.36 & 76.70 & \textbf{83.37} & 73.62 & 80.67 & 83.02 & \textcolor{NavyBlue}{+2.35} \\
         & New & 59.90 & 41.18 & 56.00 & 59.18 & 62.13 & 62.97 & 62.38 & 55.31 & \textbf{64.05} & \textcolor{NavyBlue}{+8.74} \\
         & H & 56.37 & 54.24 & 64.85 & 68.16 & 68.61 & 71.75 & 67.56 & 65.63 & \textbf{72.31} & \textcolor{NavyBlue}{+6.68} \\
         \midrule
         \multirow{3}{*}{EuroSAT} & Base & 56.48 & 92.19 & 87.49 & 94.07 & 86.63 & 92.90 & 86.26 & 92.64 & \textbf{94.94} & \textcolor{NavyBlue}{+2.30} \\
         & New & 64.05 & 54.74 & 60.04 & 73.23 & 68.97 & 73.90 & 71.38 & 63.33 & \textbf{77.44} & \textcolor{NavyBlue}{+14.11} \\
         & H & 60.03 & 68.69 & 71.21 & 82.35 & 76.79 & 82.32 & 78.12 & 75.23 & \textbf{85.30} & \textcolor{NavyBlue}{+10.07} \\
         \midrule
         \multirow{3}{*}{UCF101} & Base & 70.53 & 84.69 & 82.33 & 83.00 & 83.67 & \textbf{87.10} & 82.00 & 84.23 & 86.97 & \textcolor{NavyBlue}{+2.74} \\
         & New & 77.50 & 56.05 & 73.45 & 78.66 & 75.43 & 78.80 & 78.06 & 76.78 & \textbf{79.84} & \textcolor{NavyBlue}{+3.06} \\
         & H & 73.85 & 67.46 & 77.64 & 80.77 & 79.34 & 82.74 & 79.98 & 80.33 & \textbf{83.25} & \textcolor{NavyBlue}{+2.92} \\
    \bottomrule
    \end{tabular}
    
    }
  \vspace{-2mm}
\caption{Performance comparison on the base-to-new generalization setting.
We train our model with a subset of the classes (base classes) in a 16-shot setting and evaluate on the test set including base classes and new classes. 
H denotes the harmonic mean of base and novel performance to show the generalization trade-off~\cite{xian2017zero}.
\textbf{Avg. Rank} is the average rank of the harmonic mean on each dataset among the baselines.
$\Delta$ denotes the performance gain of ProMetaR from IVLP (our base prompting method).
}
\vspace{-5mm}
\label{tab:base2new-large}
\end{table*} 
\hjk{
We validate the effectiveness of each component of the proposed ProMetaR under the base-to-base/base-to-new setting.
Table~\ref{tab:ablations_on_components} provides the ablation study on our components, and the results are averaged over 11 datasets.
MetaLearn denotes meta-learning, TaskAug indicates task augmentation to alleviate the meta-overfitting, and MetaReg refers to meta-regularization.
Eliminating all of our components, or (a), corresponds to using only IVLP, which is the base prompt learning method of ProMetaR.
By adopting meta-learning to IVLP ((a) $\rightarrow$ (b)), 
the base class performance improves (+1.0\%) but it impairs generalization to new classes (-0.21\%).
However, our task augmentation ((b) $\rightarrow$ (c)) significantly enhances the average accuracy on new classes and harmonic mean with gains of +2.22\% and +1.48\%, respectively, compared to IVLP+meta-learning.
Additionally, our meta-regularization ((b) $\rightarrow$ (d)) improves accuracy for both base and new classes by +0.76\% and +1.91\%, respectively.
This indicates that both task augmentation and meta-regularization clearly ameliorate the meta-overfitting caused by meta-learning and contribute to strong generalization.
Furthermore, by adding meta-regularization to (c), \ie, (c) $\rightarrow$ (e), all three accuracies increase to +0.35\% (base class), +1.56\% (new class), and +1.02\% (harmonic mean).
Employing task augmentation to (d), \ie, (d) $\rightarrow$ (e), leads to an additional +1.87\% growth in new class accuracy. 
Our ProMetaR significantly improves over IVLP for both base and new classes ((a) $\rightarrow$ (e)), achieving performance gains of +1.88\%, +3.57\%, and +2.83\% on the base class, new class accuracy, and harmonic mean, respectively.
}
\subsection{Base-to-base/Base-to-new generalization}
\hjk{
We compare the performance of ProMetaR with other recent prompting approaches in the base-to-base/base-to-new generalization setting to demonstrate the effectiveness of the proposed learning framework.
Following \cite{zhou2022conditional,khattak2023maple}, we report the average accuracy of three different data splits used in CoCoOp~\cite{zhou2022conditional} for a fair comparison.
The results are reported in Table~\ref{tab:base2new-large}.

Our ProMetaR shows the best performance on the average accuracy over 11 datasets among baselines.
In particular, ProMetaR achieves a significant improvement on new classes from \textcolor{black}{76.10} to \textcolor{black}{76.93} compared to the best baseline method PromptSRC.
Also, ProMetaR substantially improves the average accuracy of the base model IVLP by \textcolor{black}{3.58} on new classes. 
This result indicates that our ProMetaR enhances the generalizability of existing prompting methods by meta-learning the regularization.
In comparison with UNIGRAM, which applies meta-learning with a large scale of external data, ProMetaR shows impressive performance improvement on both base and new categories without any external data for the meta-learning.
}
\subsection{Domain generalization}
\begin{table}[!t]
    \footnotesize 
    \centering
 \setlength{\tabcolsep}{3pt}
    \begin{tabular}{l ccccc|c}
    \toprule
    & \textbf{Source} & \multicolumn{5}{c}{\textbf{Target}} \\ \cmidrule(lr){2-2} \cmidrule(lr){3-7}
     & ImageNet & -V2 & -S & -A & -R  & Avg.\\
    \midrule
    CLIP &  66.73 & 60.83 & 46.15 & 47.77 & 73.96 & 57.18 \\
    CoOp &  71.51 & 64.20 & 47.99  & 49.71  & 75.21  & 59.28 \\
    CoCoOp & 71.02 & 64.07 & 48.75 & 50.63 & 76.18 & 59.91  \\
        MaPLe & 70.72  & 64.07 & 49.15  & 50.90 & 76.98 & 60.27 \\
        RPO & \textbf{71.67} & \textbf{65.13} & 49.27 & 50.13 & 76.57 & 60.28 \\
        PromptSRC & 71.27 & 64.35 & \textbf{49.55} & 50.90 & 77.80 & 60.65 \\
        \textcolor{Gray}{UNIGRAM} & \textcolor{Gray}{71.65} & \textcolor{Gray}{64.81} & \textcolor{Gray}{49.54} & \textcolor{Gray}{51.51} & \textcolor{Gray}{77.34} & \textcolor{Gray}{60.80} \\
    \midrule
    \rowcolor{tabhighlight} ProMetaR & 71.29 & 64.39& \textbf{49.55}& \textbf{51.25}&\textbf{77.89} & \textbf{60.77}\\
    \bottomrule
    \end{tabular}
    \vspace{-2mm}
        \caption{Performance comparison on the domain generalization.
        } 
    \label{tab:domain_generalization}
    \vspace{-4mm}
\end{table}
\hjk{
In the domain generalization setting, the performance comparison of ImageNet-trained models, evaluated with four out-of-distribution variants, is reported in Table~\ref{tab:domain_generalization}.
For a fair comparison, we exclude UNIGRAM since it employs a large scale of extra datasets to pre-train the learnable prompts.
ProMetaR successfully generalizes to out-of-domain datasets showing the best average accuracy. 
This demonstrates that our meta-regularizer and task augmentation clearly enhance the robustness to domain shifts.
}
\subsection{Analysis}

\begin{table}[t]
  \centering
  \footnotesize
  \setlength{\tabcolsep}{2pt}
  \begin{tabular}{lcc|lcc}
    \toprule
    \multicolumn{3}{c|}{Top-3} & \multicolumn{3}{c}{Bottom-3}\\
    Dataset  & $\boldsymbol{tos}^{\text{IVLP}}$ & Gain $\Delta$& Dataset  & $\boldsymbol{tos}^{\text{IVLP}}$  & Gain $\Delta$\\
    \midrule
     EuroSAT& 36.88 & 10.07 & Food101  & -0.01 & 0.25 \\
    DTD & 32.02 & 6.68  & Caltech101 & 1.79 & 0.25 \\
    Flowers & 28.25 & 3.29  & Imagenet & 3.06 & 0.56 \\
    \bottomrule
  \end{tabular}
\vspace{-2mm}
\caption{
Task overfitting score $\boldsymbol{tos}^{\text{IVLP}} =  \delta_{\text{base}}^{\text{IVLP}}-\delta_{\text{new}}^{\text{IVLP}}$ and the gain $\Delta$.
$\Delta$ denotes the performance gain (H) by ProMetaR on IVLP (Table~\ref{tab:base2new-large}).
}
\label{tab:generalizability}
\vspace{-6mm}
\end{table}


\noindent\textbf{Task overfitting score.}
We analyze when our ProMetaR provides a relatively large (or small) performance improvement compared to the base model (IVLP).
To quantify the room for improvement, 
we define Task Overfitting Score~($\boldsymbol{tos}$) of the prompting method \texttt{<pr>} as 
\begin{equation}
    \boldsymbol{tos}^{<pr>} = \delta_{\text{base}}^{\texttt{<pr>}}-\delta_{\text{new}}^{\texttt{<pr>}},
\end{equation}
where $\delta^{\texttt{<pr>}}_{\text{base}} = \max(0, s_{\text{base}}^{\texttt{<pr>}}-s_{\text{base}}^{\text{CLIP}}), \delta^{\texttt{<pr>}}_{\text{new}} = s_{\text{new}}^{\texttt{<pr>}}-s_{\text{new}}^{\text{CLIP}}$ be the performance difference between prompting method \texttt{<pr>} and zero-shot CLIP on the base and new classes, respectively.
$s_{\text{base}}^{\texttt{<pr>}}, s_{\text{new}}^{\texttt{<pr>}}$ indicate the accuracy of the prompting method \texttt{<pr>} on base and new classes, respectively.
As the task overfitting score is lower, the method \texttt{<pr>} tends to generalize well on new tasks.
Table~\ref{tab:generalizability} reports the task overfitting score and performance gain $\Delta$ of ProMetaR from IVLP (Table~\ref{tab:base2new-large}) on the datasets with top-3~(left) and bottom-3~(right) task overfitting scores.
The table shows that gains of ProMetaR are relatively high when the task overfitting score is high. 
It demonstrates that ProMetaR is more effective when prompting method IVLP suffers from overfitting.

\begin{table}[!t]
\centering
\footnotesize
{
  \begin{tabular}{l|ccc}
    \toprule
    Methods & Base & New & H\\
    \midrule
    CoOp    &   82.69 & 63.22 & 71.66 \\
    \rowcolor{tabhighlight} + ProMetaR    &  83.35 & 71.20 & 76.80 \\
    \midrule
    VPT    & 82.75  & 71.00 & 76.43 \\
    \rowcolor{tabhighlight} + ProMetaR    & 83.18  & 73.19& 77.87\\
    \bottomrule
    \end{tabular}
    }
    \vspace{-2mm}
    \caption{Performance comparison of ProMetaR with different prompting approaches~(CoOp~\cite{zhou2022learning} and VPT~\cite{jia2022visual}) under the base-to-base/base-to-new generalization setting.
    }
    \label{tab:average11}
    \vspace{-4mm}
\end{table}

\begin{table}[!t]
     \centering

 \setlength{\tabcolsep}{8pt}
    \footnotesize
    \begin{tabular}{l cc c }
    \toprule
    Method  & Base & New & H\\
    \midrule
    Loss+Reg.       & 83.96 & 75.70 & 79.62 \\
    \rowcolor{tabhighlight}
    ProMetaR~\textbf{(Ours)}  & \textbf{84.39} & \textbf{76.93} & \textbf{80.49} \\
    \rowcolor{tabhighlight}
    Performance Gain ($\Delta$)  & \textcolor{NavyBlue}{+0.43} & 
    \textcolor{NavyBlue}{+1.23} & \textcolor{NavyBlue}{+0.87} \\
    \bottomrule
    \end{tabular}
    \vspace{-2mm}
    \caption{Performance comparison of ProMetaR with IVLP trained with the loss and regularizer under the base-to-base/base-to-new generalization setting. 
    }
    \label{tab:reg_ablations}
    \vspace{-6mm}
\end{table}

\vspace{1mm}
\hjk{
\noindent\textbf{ProMetaR with diverse methods.} 
ProMetaR can be applied to any existing prompting methods in a plug-and-play manner.
We elucidate the effectiveness of ProMetaR by comparing the performance of various methods, such as CoOp and VPT, with our method plugged in (Table~\ref{tab:average11}).
ProMetaR consistently improves all the other prompt learning methods with harmonic mean gains of \textcolor{black}{+5.14\% and +1.44\%} over CoOp and VPT, respectively.
Moreover, the performance is enhanced, especially in new classes, indicating that our ProMetaR effectively prevents the prompts from overfitting to downstream tasks.

\vspace{1mm}
\noindent\textbf{Meta-Regularization.}
In Table~\ref{tab:reg_ablations}, we also compare ProMetaR with IVLP trained with the loss and the regularizer~(Loss+Reg) in \eqref{eq:loss+reg} with manually tuned hyperparameters (\eg, a regularization strength).
The experimental results show that our ProMetaR outperforms standard IVLP training with regularization (Loss+Reg). 
This result indicates that our ProMetaR automatically learns more effective regularization via meta-learning. }

\vspace{-1mm}
\section{Conclusion}
We propose ProMetaR to encourage both traditional generalization and task generalization, yielding a significant performance improvement in base-to-base/base-to-new and domain generalization settings.
Specifically, we adopt meta-learning to learn both soft prompts and regularizers.
We further incorporate task augmentation to generate diverse tasks and address the meta-overfitting.
Extensive experiments and analyses demonstrate that our ProMetaR enhances the generalizability of prompt learning.

\noindent\textbf{Acknowledgements.}
This work was partly supported by ICT Creative Consilience Program through the IITP, NRF of Korea grants funded by the Korea government (MSIT) (IITP-2024-2020-0-01819, NRF-2023R1A2C2005373), and the NVIDIA academic grant.
We thank Jongha Kim for the suggestions on the analysis.

{
    \small
    \bibliographystyle{ieeenat_fullname}
    \bibliography{main}

\begin{thebibliography}{85}
\providecommand{\natexlab}[1]{#1}
\providecommand{\url}[1]{\texttt{#1}}
\expandafter\ifx\csname urlstyle\endcsname\relax
  \providecommand{\doi}[1]{doi: #1}\else
  \providecommand{\doi}{doi: \begingroup \urlstyle{rm}\Url}\fi

\bibitem[Allingham et~al.(2023)Allingham, Ren, Dusenberry, Gu, Cui, Tran, Liu, and Lakshminarayanan]{allingham2023simple}
James~Urquhart Allingham, Jie Ren, Michael~W Dusenberry, Xiuye Gu, Yin Cui, Dustin Tran, Jeremiah~Zhe Liu, and Balaji Lakshminarayanan.
\newblock A simple zero-shot prompt weighting technique to improve prompt ensembling in text-image models.
\newblock In \emph{ICML}, 2023.

\bibitem[Antoniou et~al.(2019)Antoniou, Edwards, and Storkey]{antoniou2018train}
Antreas Antoniou, Harrison Edwards, and Amos Storkey.
\newblock How to train your maml.
\newblock In \emph{ICLR}, 2019.

\bibitem[Balaji et~al.(2018)Balaji, Sankaranarayanan, and Chellappa]{DBLP:conf/nips/BalajiSC18}
Yogesh Balaji, Swami Sankaranarayanan, and Rama Chellappa.
\newblock Metareg: Towards domain generalization using meta-regularization.
\newblock In \emph{NeurIPS}, 2018.

\bibitem[Bechtle et~al.(2020)Bechtle, Molchanov, Chebotar, Grefenstette, Righetti, Sukhatme, and Meier]{DBLP:conf/icpr/BechtleMCGRSM20}
Sarah Bechtle, Artem Molchanov, Yevgen Chebotar, Edward Grefenstette, Ludovic Righetti, Gaurav~S. Sukhatme, and Franziska Meier.
\newblock Meta learning via learned loss.
\newblock In \emph{ICPR}, 2020.

\bibitem[Bossard et~al.(2014)Bossard, Guillaumin, and Van~Gool]{bossard2014food}
Lukas Bossard, Matthieu Guillaumin, and Luc Van~Gool.
\newblock Food-101--mining discriminative components with random forests.
\newblock In \emph{ECCV}, 2014.

\bibitem[Cho et~al.(2023)Cho, Kim, and Kim]{cho2023distribution}
Eulrang Cho, Jooyeon Kim, and Hyunwoo~J Kim.
\newblock Distribution-aware prompt tuning for vision-language models.
\newblock In \emph{ICCV}, 2023.

\bibitem[Choi et~al.(2023)Choi, Choi, and Kim]{choi2023tokenmixup}
Hyeong~Kyu Choi, Joonmyung Choi, and Hyunwoo~J Kim.
\newblock Tokenmixup: Efficient attention-guided token-level data augmentation for transformers.
\newblock In \emph{NeurIPS}, 2023.

\bibitem[Cimpoi et~al.(2014)Cimpoi, Maji, Kokkinos, Mohamed, and Vedaldi]{cimpoi2014describing}
Mircea Cimpoi, Subhransu Maji, Iasonas Kokkinos, Sammy Mohamed, and Andrea Vedaldi.
\newblock Describing textures in the wild.
\newblock In \emph{CVPR}, 2014.

\bibitem[Deng et~al.(2009)Deng, Dong, Socher, Li, Li, and Fei-Fei]{deng2009imagenet}
Jia Deng, Wei Dong, Richard Socher, Li-Jia Li, Kai Li, and Li Fei-Fei.
\newblock Imagenet: A large-scale hierarchical image database.
\newblock In \emph{CVPR}, 2009.

\bibitem[Dosovitskiy et~al.(2021)Dosovitskiy, Beyer, Kolesnikov, Weissenborn, Zhai, Unterthiner, Dehghani, Minderer, Heigold, Gelly, et~al.]{dosovitskiy2021image}
Alexey Dosovitskiy, Lucas Beyer, Alexander Kolesnikov, Dirk Weissenborn, Xiaohua Zhai, Thomas Unterthiner, Mostafa Dehghani, Matthias Minderer, Georg Heigold, Sylvain Gelly, et~al.
\newblock An image is worth 16x16 words: Transformers for image recognition at scale.
\newblock In \emph{ICLR}, 2021.

\bibitem[Du et~al.(2022)Du, Wei, Zhang, Shi, Gao, and Li]{du2022learning}
Yu Du, Fangyun Wei, Zihe Zhang, Miaojing Shi, Yue Gao, and Guoqi Li.
\newblock Learning to prompt for open-vocabulary object detection with vision-language model.
\newblock In \emph{CVPR}, 2022.

\bibitem[Fei-Fei et~al.(2004)Fei-Fei, Fergus, and Perona]{fei2004learning}
Li Fei-Fei, Rob Fergus, and Pietro Perona.
\newblock Learning generative visual models from few training examples: An incremental bayesian approach tested on 101 object categories.
\newblock In \emph{CVPRW}, 2004.

\bibitem[Feng et~al.(2022)Feng, Zhong, Jie, Chu, Ren, Wei, Xie, and Ma]{feng2022promptdet}
Chengjian Feng, Yujie Zhong, Zequn Jie, Xiangxiang Chu, Haibing Ren, Xiaolin Wei, Weidi Xie, and Lin Ma.
\newblock Promptdet: Towards open-vocabulary detection using uncurated images.
\newblock In \emph{ECCV}, 2022.

\bibitem[Finn et~al.(2017)Finn, Abbeel, and Levine]{finn2017model}
Chelsea Finn, Pieter Abbeel, and Sergey Levine.
\newblock Model-agnostic meta-learning for fast adaptation of deep networks.
\newblock In \emph{ICML}, 2017.

\bibitem[Grant et~al.(2018)Grant, Finn, Levine, Darrell, and Griffiths]{grant2018recasting}
Erin Grant, Chelsea Finn, Sergey Levine, Trevor Darrell, and Thomas Griffiths.
\newblock Recasting gradient-based meta-learning as hierarchical bayes.
\newblock In \emph{ICLR}, 2018.

\bibitem[Gu et~al.(2022)Gu, Lin, Kuo, and Cui]{gu2022open}
Xiuye Gu, Tsung-Yi Lin, Weicheng Kuo, and Yin Cui.
\newblock Open-vocabulary object detection via vision and language knowledge distillation.
\newblock In \emph{ICLR}, 2022.

\bibitem[Helber et~al.(2019)Helber, Bischke, Dengel, and Borth]{helber2019eurosat}
Patrick Helber, Benjamin Bischke, Andreas Dengel, and Damian Borth.
\newblock Eurosat: A novel dataset and deep learning benchmark for land use and land cover classification.
\newblock \emph{JSTARS}, 12\penalty0 (7):\penalty0 2217--2226, 2019.

\bibitem[Hendrycks et~al.(2021{\natexlab{a}})Hendrycks, Basart, Mu, Kadavath, Wang, Dorundo, Desai, Zhu, Parajuli, Guo, et~al.]{hendrycks2021many}
Dan Hendrycks, Steven Basart, Norman Mu, Saurav Kadavath, Frank Wang, Evan Dorundo, Rahul Desai, Tyler Zhu, Samyak Parajuli, Mike Guo, et~al.
\newblock The many faces of robustness: A critical analysis of out-of-distribution generalization.
\newblock In \emph{ICCV}, 2021{\natexlab{a}}.

\bibitem[Hendrycks et~al.(2021{\natexlab{b}})Hendrycks, Zhao, Basart, Steinhardt, and Song]{hendrycks2021natural}
Dan Hendrycks, Kevin Zhao, Steven Basart, Jacob Steinhardt, and Dawn Song.
\newblock Natural adversarial examples.
\newblock In \emph{CVPR}, 2021{\natexlab{b}}.

\bibitem[Hochreiter et~al.(2001)Hochreiter, Younger, and Conwell]{hochreiter2001learning}
Sepp Hochreiter, A~Steven Younger, and Peter~R Conwell.
\newblock Learning to learn using gradient descent.
\newblock In \emph{ICANN}, 2001.

\bibitem[Hospedales et~al.(2021)Hospedales, Antoniou, Micaelli, and Storkey]{hospedales2021meta}
Timothy Hospedales, Antreas Antoniou, Paul Micaelli, and Amos Storkey.
\newblock Meta-learning in neural networks: A survey.
\newblock \emph{TPAMI}, 44\penalty0 (9):\penalty0 5149--5169, 2021.

\bibitem[Hwang et~al.(2020)Hwang, Park, Kwon, Kim, Ha, and Kim]{hwang2020self}
Dasol Hwang, Jinyoung Park, Sunyoung Kwon, KyungMin Kim, Jung-Woo Ha, and Hyunwoo~J Kim.
\newblock Self-supervised auxiliary learning with meta-paths for heterogeneous graphs.
\newblock In \emph{NeurIPS}, 2020.

\bibitem[Hwang et~al.(2021)Hwang, Park, Kwon, Kim, Ha, and Kim]{hwang2021self}
Dasol Hwang, Jinyoung Park, Sunyoung Kwon, Kyung-Min Kim, Jung-Woo Ha, and Hyunwoo~J Kim.
\newblock Self-supervised auxiliary learning for graph neural networks via meta-learning.
\newblock \emph{arXiv:2103.00771}, 2021.

\bibitem[Ilharco et~al.(2022)Ilharco, Wortsman, Gadre, Song, Hajishirzi, Kornblith, Farhadi, and Schmidt]{ilharco2022patching}
Gabriel Ilharco, Mitchell Wortsman, Samir~Yitzhak Gadre, Shuran Song, Hannaneh Hajishirzi, Simon Kornblith, Ali Farhadi, and Ludwig Schmidt.
\newblock Patching open-vocabulary models by interpolating weights.
\newblock In \emph{NeurIPS}, 2022.

\bibitem[Izmailov et~al.(2018)Izmailov, Podoprikhin, Garipov, Vetrov, and Wilson]{izmailov2018averaging}
Pavel Izmailov, Dmitrii Podoprikhin, Timur Garipov, Dmitry Vetrov, and Andrew~Gordon Wilson.
\newblock Averaging weights leads to wider optima and better generalization.
\newblock In \emph{UAI}, 2018.

\bibitem[Jia et~al.(2021)Jia, Yang, Xia, Chen, Parekh, Pham, Le, Sung, Li, and Duerig]{jia2021scaling}
Chao Jia, Yinfei Yang, Ye Xia, Yi-Ting Chen, Zarana Parekh, Hieu Pham, Quoc Le, Yun-Hsuan Sung, Zhen Li, and Tom Duerig.
\newblock Scaling up visual and vision-language representation learning with noisy text supervision.
\newblock In \emph{ICML}, 2021.

\bibitem[Jia et~al.(2022)Jia, Tang, Chen, Cardie, Belongie, Hariharan, and Lim]{jia2022visual}
Menglin Jia, Luming Tang, Bor-Chun Chen, Claire Cardie, Serge Belongie, Bharath Hariharan, and Ser-Nam Lim.
\newblock Visual prompt tuning.
\newblock In \emph{ECCV}, 2022.

\bibitem[Khattak et~al.(2023{\natexlab{a}})Khattak, Rasheed, Maaz, Khan, and Khan]{khattak2023maple}
Muhammad~Uzair Khattak, Hanoona Rasheed, Muhammad Maaz, Salman Khan, and Fahad~Shahbaz Khan.
\newblock Maple: Multi-modal prompt learning.
\newblock In \emph{CVPR}, 2023{\natexlab{a}}.

\bibitem[Khattak et~al.(2023{\natexlab{b}})Khattak, Wasim, Naseer, Khan, Yang, and Khan]{khattak2023self}
Muhammad~Uzair Khattak, Syed~Talal Wasim, Muzammal Naseer, Salman Khan, Ming-Hsuan Yang, and Fahad~Shahbaz Khan.
\newblock Self-regulating prompts: Foundational model adaptation without forgetting.
\newblock In \emph{ICCV}, 2023{\natexlab{b}}.

\bibitem[Kim et~al.(2021)Kim, Choo, Jeong, and Song]{kim2021co}
Jang-Hyun Kim, Wonho Choo, Hosan Jeong, and Hyun~Oh Song.
\newblock Co-mixup: Saliency guided joint mixup with supermodular diversity.
\newblock In \emph{ICLR}, 2021.

\bibitem[Ko et~al.(2023)Ko, Choi, Choi, On, Roh, and Kim]{ko2023meltr}
Dohwan Ko, Joonmyung Choi, Hyeong~Kyu Choi, Kyoung-Woon On, Byungseok Roh, and Hyunwoo~J Kim.
\newblock Meltr: Meta loss transformer for learning to fine-tune video foundation models.
\newblock In \emph{CVPR}, 2023.

\bibitem[Koch et~al.(2015)Koch, Zemel, Salakhutdinov, et~al.]{koch2015siamese}
Gregory Koch, Richard Zemel, Ruslan Salakhutdinov, et~al.
\newblock Siamese neural networks for one-shot image recognition.
\newblock In \emph{ICMLW}, 2015.

\bibitem[Krause et~al.(2013)Krause, Stark, Deng, and Fei-Fei]{krause20133d}
Jonathan Krause, Michael Stark, Jia Deng, and Li Fei-Fei.
\newblock 3d object representations for fine-grained categorization.
\newblock In \emph{ICCVW}, 2013.

\bibitem[Lee et~al.(2023)Lee, Song, Suh, Choi, Lee, and Kim]{lee2023read}
Dongjun Lee, Seokwon Song, Jihee Suh, Joonmyeong Choi, Sanghyeok Lee, and Hyunwoo~J Kim.
\newblock Read-only prompt optimization for vision-language few-shot learning.
\newblock In \emph{ICCV}, 2023.

\bibitem[Lee et~al.(2019)Lee, Maji, Ravichandran, and Soatto]{lee2019meta}
Kwonjoon Lee, Subhransu Maji, Avinash Ravichandran, and Stefano Soatto.
\newblock Meta-learning with differentiable convex optimization.
\newblock In \emph{CVPR}, 2019.

\bibitem[Lester et~al.(2021)Lester, Al-Rfou, and Constant]{lester2021power}
Brian Lester, Rami Al-Rfou, and Noah Constant.
\newblock The power of scale for parameter-efficient prompt tuning.
\newblock In \emph{EMNLP}, 2021.

\bibitem[Li et~al.(2018)Li, Yang, Song, and Hospedales]{li2018learning}
Da Li, Yongxin Yang, Yi-Zhe Song, and Timothy Hospedales.
\newblock Learning to generalize: Meta-learning for domain generalization.
\newblock In \emph{AAAI}, 2018.

\bibitem[Li et~al.(2023{\natexlab{a}})Li, Gao, Wei, Tang, Zhang, Li, Ji, Tian, Chua, and Zhuang]{li2023gradient}
Juncheng Li, Minghe Gao, Longhui Wei, Siliang Tang, Wenqiao Zhang, Mengze Li, Wei Ji, Qi Tian, Tat-Seng Chua, and Yueting Zhuang.
\newblock Gradient-regulated meta-prompt learning for generalizable vision-language models.
\newblock In \emph{ICCV}, 2023{\natexlab{a}}.

\bibitem[Li et~al.(2023{\natexlab{b}})Li, Li, Savarese, and Hoi]{li2023blip}
Junnan Li, Dongxu Li, Silvio Savarese, and Steven Hoi.
\newblock Blip-2: Bootstrapping language-image pre-training with frozen image encoders and large language models.
\newblock In \emph{ICML}, 2023{\natexlab{b}}.

\bibitem[Li and Liang(2021)]{li2021prefix}
Xiang~Lisa Li and Percy Liang.
\newblock Prefix-tuning: Optimizing continuous prompts for generation.
\newblock In \emph{ACL}, 2021.

\bibitem[Liu et~al.(2023)Liu, Zheng, Du, Ding, Qian, Yang, and Tang]{liu2023gpt}
Xiao Liu, Yanan Zheng, Zhengxiao Du, Ming Ding, Yujie Qian, Zhilin Yang, and Jie Tang.
\newblock Gpt understands, too.
\newblock \emph{AI Open}, 2023.

\bibitem[Loshchilov and Hutter(2019)]{loshchilov2019decoupled}
Ilya Loshchilov and Frank Hutter.
\newblock Decoupled weight decay regularization.
\newblock In \emph{ICLR}, 2019.

\bibitem[Lu et~al.(2022)Lu, Liu, Zhang, Liu, and Tian]{lu2022prompt}
Yuning Lu, Jianzhuang Liu, Yonggang Zhang, Yajing Liu, and Xinmei Tian.
\newblock Prompt distribution learning.
\newblock In \emph{CVPR}, 2022.

\bibitem[L{\"u}ddecke and Ecker(2022)]{luddecke2022image}
Timo L{\"u}ddecke and Alexander Ecker.
\newblock Image segmentation using text and image prompts.
\newblock In \emph{CVPR}, 2022.

\bibitem[Maji et~al.(2013)Maji, Kannala, Rahtu, Blaschko, and Vedaldi]{maji2013fine}
S. Maji, J. Kannala, E. Rahtu, M. Blaschko, and A. Vedaldi.
\newblock Fine-grained visual classification of aircraft.
\newblock Technical report, 2013.

\bibitem[Mishra et~al.(2018)Mishra, Rohaninejad, Chen, and Abbeel]{mishra2018simple}
Nikhil Mishra, Mostafa Rohaninejad, Xi Chen, and Pieter Abbeel.
\newblock A simple neural attentive meta-learner.
\newblock In \emph{ICLR}, 2018.

\bibitem[Mokady et~al.(2021)Mokady, Hertz, and Bermano]{mokady2021clipcap}
Ron Mokady, Amir Hertz, and Amit~H Bermano.
\newblock Clipcap: Clip prefix for image captioning.
\newblock \emph{arXiv:2111.09734}, 2021.

\bibitem[Munkhdalai and Yu(2017)]{munkhdalai2017meta}
Tsendsuren Munkhdalai and Hong Yu.
\newblock Meta networks.
\newblock In \emph{ICML}, 2017.

\bibitem[Munkhdalai et~al.(2018)Munkhdalai, Yuan, Mehri, and Trischler]{munkhdalai2018rapid}
Tsendsuren Munkhdalai, Xingdi Yuan, Soroush Mehri, and Adam Trischler.
\newblock Rapid adaptation with conditionally shifted neurons.
\newblock In \emph{ICML}, 2018.

\bibitem[Nichol et~al.(2018)Nichol, Achiam, and Schulman]{nichol2018first}
Alex Nichol, Joshua Achiam, and John Schulman.
\newblock On first-order meta-learning algorithms.
\newblock \emph{arXiv:1803.02999}, 2018.

\bibitem[Nilsback and Zisserman(2008)]{nilsback2008automated}
Maria-Elena Nilsback and Andrew Zisserman.
\newblock Automated flower classification over a large number of classes.
\newblock In \emph{ICVGIP}, 2008.

\bibitem[Park et~al.(2022)Park, Lee, Kim, Park, Jeong, Kim, Ha, and Kim]{park2021metropolis}
Hyeonjin Park, Seunghun Lee, Sihyeon Kim, Jinyoung Park, Jisu Jeong, Kyung-Min Kim, Jung-Woo Ha, and Hyunwoo~J Kim.
\newblock Metropolis-hastings data augmentation for graph neural networks.
\newblock In \emph{NeurIPS}, 2022.

\bibitem[Parkhi et~al.(2012)Parkhi, Vedaldi, Zisserman, and Jawahar]{parkhi2012cats}
Omkar~M Parkhi, Andrea Vedaldi, Andrew Zisserman, and CV Jawahar.
\newblock Cats and dogs.
\newblock In \emph{CVPR}, 2012.

\bibitem[Paszke et~al.(2017)Paszke, Gross, Chintala, Chanan, Yang, DeVito, Lin, Desmaison, Antiga, and Lerer]{paszke2017automatic}
Adam Paszke, Sam Gross, Soumith Chintala, Gregory Chanan, Edward Yang, Zachary DeVito, Zeming Lin, Alban Desmaison, Luca Antiga, and Adam Lerer.
\newblock Automatic differentiation in pytorch.
\newblock In \emph{ICLRW}, 2017.

\bibitem[Radford et~al.(2021)Radford, Kim, Hallacy, Ramesh, Goh, Agarwal, Sastry, Askell, Mishkin, Clark, et~al.]{radford2021learning}
Alec Radford, Jong~Wook Kim, Chris Hallacy, Aditya Ramesh, Gabriel Goh, Sandhini Agarwal, Girish Sastry, Amanda Askell, Pamela Mishkin, Jack Clark, et~al.
\newblock Learning transferable visual models from natural language supervision.
\newblock In \emph{ICML}, 2021.

\bibitem[Ravi and Larochelle(2016)]{ravi2016optimization}
Sachin Ravi and Hugo Larochelle.
\newblock Optimization as a model for few-shot learning.
\newblock In \emph{ICLR}, 2016.

\bibitem[Recht et~al.(2019)Recht, Roelofs, Schmidt, and Shankar]{recht2019imagenet}
Benjamin Recht, Rebecca Roelofs, Ludwig Schmidt, and Vaishaal Shankar.
\newblock Do imagenet classifiers generalize to imagenet?
\newblock In \emph{ICML}, 2019.

\bibitem[Santoro et~al.(2016)Santoro, Bartunov, Botvinick, Wierstra, and Lillicrap]{santoro2016meta}
Adam Santoro, Sergey Bartunov, Matthew Botvinick, Daan Wierstra, and Timothy Lillicrap.
\newblock Meta-learning with memory-augmented neural networks.
\newblock In \emph{ICML}, 2016.

\bibitem[Shu et~al.(2019)Shu, Xie, Yi, Zhao, Zhou, Xu, and Meng]{DBLP:conf/nips/ShuXY0ZXM19}
Jun Shu, Qi Xie, Lixuan Yi, Qian Zhao, Sanping Zhou, Zongben Xu, and Deyu Meng.
\newblock Meta-weight-net: Learning an explicit mapping for sample weighting.
\newblock In \emph{NeurIPS}, 2019.

\bibitem[Singh et~al.(2022)Singh, Hu, Goswami, Couairon, Galuba, Rohrbach, and Kiela]{singh2022flava}
Amanpreet Singh, Ronghang Hu, Vedanuj Goswami, Guillaume Couairon, Wojciech Galuba, Marcus Rohrbach, and Douwe Kiela.
\newblock Flava: A foundational language and vision alignment model.
\newblock In \emph{CVPR}, 2022.

\bibitem[Snell et~al.(2017)Snell, Swersky, and Zemel]{snell2017prototypical}
Jake Snell, Kevin Swersky, and Richard Zemel.
\newblock Prototypical networks for few-shot learning.
\newblock In \emph{NeurIPS}, 2017.

\bibitem[Soomro et~al.(2013)Soomro, Zamir, and Shah]{soomro2012ucf101}
Khurram Soomro, Amir~Roshan Zamir, and Mubarak Shah.
\newblock Ucf101: A dataset of 101 human actions classes from videos in the wild.
\newblock In \emph{ICCVW}, 2013.

\bibitem[Srivastava et~al.(2014)Srivastava, Hinton, Krizhevsky, Sutskever, and Salakhutdinov]{srivastava2014dropout}
Nitish Srivastava, Geoffrey Hinton, Alex Krizhevsky, Ilya Sutskever, and Ruslan Salakhutdinov.
\newblock Dropout: a simple way to prevent neural networks from overfitting.
\newblock \emph{JMLR}, 15\penalty0 (1):\penalty0 1929--1958, 2014.

\bibitem[Sung et~al.(2018)Sung, Yang, Zhang, Xiang, Torr, and Hospedales]{sung2018learning}
Flood Sung, Yongxin Yang, Li Zhang, Tao Xiang, Philip~HS Torr, and Timothy~M Hospedales.
\newblock Learning to compare: Relation network for few-shot learning.
\newblock In \emph{CVPR}, 2018.

\bibitem[Uddin et~al.(2021)Uddin, Monira, Shin, Chung, Bae, et~al.]{uddin2021saliencymix}
AFM Uddin, Mst Monira, Wheemyung Shin, TaeChoong Chung, Sung-Ho Bae, et~al.
\newblock Saliencymix: A saliency guided data augmentation strategy for better regularization.
\newblock In \emph{ICLR}, 2021.

\bibitem[Verma et~al.(2019)Verma, Lamb, Beckham, Najafi, Courville, Mitliagkas, and Bengio]{verma2018manifold}
Vikas Verma, Alex Lamb, Christopher Beckham, Amir Najafi, Aaron Courville, Ioannis Mitliagkas, and Yoshua Bengio.
\newblock Manifold mixup: learning better representations by interpolating hidden states.
\newblock In \emph{ICML}, 2019.

\bibitem[Vinyals et~al.(2016)Vinyals, Blundell, Lillicrap, Wierstra, et~al.]{vinyals2016matching}
Oriol Vinyals, Charles Blundell, Timothy Lillicrap, Daan Wierstra, et~al.
\newblock Matching networks for one shot learning.
\newblock In \emph{NeurIPS}, 2016.

\bibitem[Wang et~al.(2019)Wang, Ge, Lipton, and Xing]{wang2019learning}
Haohan Wang, Songwei Ge, Zachary Lipton, and Eric~P Xing.
\newblock Learning robust global representations by penalizing local predictive power.
\newblock In \emph{NeurIPS}, 2019.

\bibitem[Wortsman et~al.(2022)Wortsman, Ilharco, Kim, Li, Kornblith, Roelofs, Lopes, Hajishirzi, Farhadi, Namkoong, et~al.]{wortsman2022robust}
Mitchell Wortsman, Gabriel Ilharco, Jong~Wook Kim, Mike Li, Simon Kornblith, Rebecca Roelofs, Raphael~Gontijo Lopes, Hannaneh Hajishirzi, Ali Farhadi, Hongseok Namkoong, et~al.
\newblock Robust fine-tuning of zero-shot models.
\newblock In \emph{CVPR}, 2022.

\bibitem[Xian et~al.(2017)Xian, Schiele, and Akata]{xian2017zero}
Yongqin Xian, Bernt Schiele, and Zeynep Akata.
\newblock Zero-shot learning-the good, the bad and the ugly.
\newblock In \emph{CVPR}, 2017.

\bibitem[Xiao et~al.(2010)Xiao, Hays, Ehinger, Oliva, and Torralba]{xiao2010sun}
Jianxiong Xiao, James Hays, Krista~A Ehinger, Aude Oliva, and Antonio Torralba.
\newblock Sun database: Large-scale scene recognition from abbey to zoo.
\newblock In \emph{CVPR}, 2010.

\bibitem[Yao et~al.(2022)Yao, Zhang, and Finn]{yao2021meta}
Huaxiu Yao, Linjun Zhang, and Chelsea Finn.
\newblock Meta-learning with fewer tasks through task interpolation.
\newblock In \emph{ICLR}, 2022.

\bibitem[Yun et~al.(2019)Yun, Han, Oh, Chun, Choe, and Yoo]{yun2019cutmix}
Sangdoo Yun, Dongyoon Han, Seong~Joon Oh, Sanghyuk Chun, Junsuk Choe, and Youngjoon Yoo.
\newblock Cutmix: Regularization strategy to train strong classifiers with localizable features.
\newblock In \emph{ICCV}, 2019.

\bibitem[Zang et~al.(2022)Zang, Li, Zhou, Huang, and Loy]{zang2022unified}
Yuhang Zang, Wei Li, Kaiyang Zhou, Chen Huang, and Chen~Change Loy.
\newblock Unified vision and language prompt learning.
\newblock \emph{arXiv:2210.07225}, 2022.

\bibitem[Zhai et~al.(2022)Zhai, Wang, Mustafa, Steiner, Keysers, Kolesnikov, and Beyer]{zhai2022lit}
Xiaohua Zhai, Xiao Wang, Basil Mustafa, Andreas Steiner, Daniel Keysers, Alexander Kolesnikov, and Lucas Beyer.
\newblock Lit: Zero-shot transfer with locked-image text tuning.
\newblock In \emph{CVPR}, 2022.

\bibitem[Zhang et~al.(2019)Zhang, Wang, Xu, and Grosse]{zhang2019three}
Guodong Zhang, Chaoqi Wang, Bowen Xu, and Roger Grosse.
\newblock Three mechanisms of weight decay regularization.
\newblock In \emph{ICLR}, 2019.

\bibitem[Zhang et~al.(2018)Zhang, Cisse, Dauphin, and Lopez-Paz]{zhang2017mixup}
Hongyi Zhang, Moustapha Cisse, Yann~N Dauphin, and David Lopez-Paz.
\newblock mixup: Beyond empirical risk minimization.
\newblock In \emph{ICLR}, 2018.

\bibitem[Zhang et~al.(2024)Zhang, Han, Zhou, Hu, Yan, Lu, Li, Gao, and Qiao]{zhang2023llama}
Renrui Zhang, Jiaming Han, Aojun Zhou, Xiangfei Hu, Shilin Yan, Pan Lu, Hongsheng Li, Peng Gao, and Yu Qiao.
\newblock Llama-adapter: Efficient fine-tuning of language models with zero-init attention.
\newblock In \emph{ICLR}, 2024.

\bibitem[Zhong et~al.(2022)Zhong, Yang, Zhang, Li, Codella, Li, Zhou, Dai, Yuan, Li, et~al.]{zhong2022regionclip}
Yiwu Zhong, Jianwei Yang, Pengchuan Zhang, Chunyuan Li, Noel Codella, Liunian~Harold Li, Luowei Zhou, Xiyang Dai, Lu Yuan, Yin Li, et~al.
\newblock Regionclip: Region-based language-image pretraining.
\newblock In \emph{CVPR}, 2022.

\bibitem[Zhou et~al.(2021)Zhou, Yang, Qiao, and Xiang]{zhou2021domain}
Kaiyang Zhou, Yongxin Yang, Yu Qiao, and Tao Xiang.
\newblock Domain adaptive ensemble learning.
\newblock \emph{TIP}, 30:\penalty0 8008--8018, 2021.

\bibitem[Zhou et~al.(2022{\natexlab{a}})Zhou, Liu, Qiao, Xiang, and Loy]{zhou2022domain}
Kaiyang Zhou, Ziwei Liu, Yu Qiao, Tao Xiang, and Chen~Change Loy.
\newblock Domain generalization: A survey.
\newblock \emph{TPAMI}, 2022{\natexlab{a}}.

\bibitem[Zhou et~al.(2022{\natexlab{b}})Zhou, Yang, Loy, and Liu]{zhou2022conditional}
Kaiyang Zhou, Jingkang Yang, Chen~Change Loy, and Ziwei Liu.
\newblock Conditional prompt learning for vision-language models.
\newblock In \emph{CVPR}, 2022{\natexlab{b}}.

\bibitem[Zhou et~al.(2022{\natexlab{c}})Zhou, Yang, Loy, and Liu]{zhou2022learning}
Kaiyang Zhou, Jingkang Yang, Chen~Change Loy, and Ziwei Liu.
\newblock Learning to prompt for vision-language models.
\newblock \emph{IJCV}, 130\penalty0 (9):\penalty0 2337--2348, 2022{\natexlab{c}}.

\bibitem[Zhu et~al.(2023)Zhu, Niu, Han, Wu, and Zhang]{zhu2023prompt}
Beier Zhu, Yulei Niu, Yucheng Han, Yue Wu, and Hanwang Zhang.
\newblock Prompt-aligned gradient for prompt tuning.
\newblock In \emph{ICCV}, 2023.

\bibitem[Zintgraf et~al.(2019)Zintgraf, Shiarli, Kurin, Hofmann, and Whiteson]{zintgraf2019fast}
Luisa Zintgraf, Kyriacos Shiarli, Vitaly Kurin, Katja Hofmann, and Shimon Whiteson.
\newblock Fast context adaptation via meta-learning.
\newblock In \emph{ICML}, 2019.

\end{thebibliography}
}

\clearpage
\appendix
In this supplement, we provide the implementation details~(Section~\ref{sec:a}) and additional experimental results~(Section~\ref{sec:b}).
\section{Implementation details}
\label{sec:a}
In this section, we provide the implementation details of our work.
We implement our ProMetaR~(\textbf{Pro}mpt learning via \textbf{Meta} \textbf{R}egularization) using Pytorch~\cite{paszke2017automatic} and Dassl~\cite{zhou2021domain,zhou2022domain}, which is a library designed for domain adaptation and generalization.
Following previous prompt learning methods~\cite{zhou2022conditional,khattak2023self,khattak2023maple}, we use CLIP-ViT-B/16 as the pretrained backbone model~\cite{radford2021learning} and four soft prompting tokens for each modality.
Following other works~\cite{allingham2023simple,radford2021learning,khattak2023self}, we utilize an ensemble of text prompts for the textual regularizer.
For the base prompt learning method, we use Independent Vision-Language Prompting as a base prompt learning method that optimizes hierarchical prompts on both image and text modalities~\cite{khattak2023maple}. 
The learning rate is set to 0.0025, and the prompts are optimized with SGD optimizer for all experiments.
For the base-to-new generalization settings, we train the model for 15 epochs. 
For domain generalization and cross-dataset transfer settings, we train the models for 6 epochs.
In all experiments, we evaluate the performance of the methods in three independent runs (seed 1, 2, and 3) and report average performance following previous prompt learning approaches~\cite{zhou2022conditional,khattak2023self,khattak2023maple}.

\paragraph{Evaluation metrics.}
In all experiments, we report top-1 accuracy for each dataset.
In base-to-novel generalization, the top-1 accuracy is measured on base classes and new classes, respectively.
We calculate the harmonic mean (H) between the base and new class accuracy to show the generalization trade-off~\cite{xian2017zero}.
In domain generalization, and cross-dataset evaluation settings, we measure top-1 accuracy on the test set of each dataset with the split provided by CoOp~\cite{zhou2022learning} following other prompt optimization works.

\section{Additional experiments}
\label{sec:b}
In this section, we provide the results of the additional experiments including cross-dataset settings and more analysis.

\subsection{Cross-dataset}
We also measure the performance of the proposed method in the cross-dataset transfer setting to explore the \textit{task} generalization ability of ProMetaR in Table~\ref{tab:cross-dataset}.
In cross-dataset transfer setting, we train our ProMetaR on ImageNet~\cite{deng2009imagenet} as a source dataset and evaluate it on other 11 \textit{unseen} datasets such as Caltech101, OxfordPets, StanfordCars, Flowers102, Food101, FGVC Aircraft, Sun397, DTD, EuroSAT, and UCF101 following other works.
Please note that the model cannot access the \textit{unseen} datasets during the training phase.
\begin{table*}[!t]
    \tabstyle{3.0pt}
    {
    \begin{tabular}{l c cccccccccc}
    \toprule
    & \textbf{Source} & \multicolumn{10}{c}{\textbf{Target}} \\ \cmidrule(lr){2-2} \cmidrule(lr){3-12}
    & \rotatebox{90}{ImageNet} & \rotatebox{90}{Caltech101} & \rotatebox{90}{OxfordPets} & \rotatebox{90}{StanfordCars} & \rotatebox{90}{Flowers102} & \rotatebox{90}{Food101} & \rotatebox{90}{Aircraft} & \rotatebox{90}{SUN397} & \rotatebox{90}{DTD} & \rotatebox{90}{EuroSAT} & \rotatebox{90}{UCF101} \\
    \midrule
    CoOp & 71.51 & 93.70 & 89.14 & 64.51 & 68.71 & 85.30 & 18.47 & 64.15 & 41.92 & 46.39 & 66.55\\
    CoCoOp & 71.02 & \textbf{94.43} & 90.14 & 65.32 & 71.88 & 86.06 & 22.94 & 67.36 & 45.73 & 45.37 & 68.21 \\
    MaPLe & 70.72 & 93.53 & 90.49 & 65.57 & \textbf{72.23} & 86.20 & 24.74 & 67.01 & 46.49 & \textbf{48.06} & 68.69 \\
    PromptSRC & 71.27 & 93.60 & 90.25 & 65.70 & 70.25 & 86.15 & 23.90 & 67.10 & 46.87 & 45.50 & 68.75 \\
    \textcolor{Gray}{UNIGRAM} & \textcolor{Gray}{71.65} & \textcolor{Gray}{94.67} & \textcolor{Gray}{90.83} & \textcolor{Gray}{66.78} & \textcolor{Gray}{73.12} & \textcolor{Gray}{86.69} & \textcolor{Gray}{25.27} & \textcolor{Gray}{67.97} & \textcolor{Gray}{48.06} & \textcolor{Gray}{52.63} & \textcolor{Gray}{71.03}  \\
    \midrule
\rowcolor{tabhighlight} ProMetaR & \textbf{71.29} & 93.74 & \textbf{90.59} & \textbf{65.83} & 71.13 & \textbf{86.39} & \textbf{24.78} & \textbf{67.41} & \textbf{47.08} & 45.02 & \textbf{69.50}\\
    \bottomrule
    \end{tabular}
    }
        \caption{Performance comparison on the cross-dataset transfer setting.}
    \label{tab:cross-dataset}
\end{table*}

For a fair comparison, we exclude UNIGRAM since it employs a large scale of extra datasets to pre-train the learnable prompts.
From the table, ProMetaR successfully generalizes on out-of-domain datasets, achieving the best performance on 7 out of 10 datasets compared to other baselines. 
This result indicates that our ProMetaR improves the \textit{task} generalization ability of the existing prompting methods and robustness against domain shifts.

\subsection{More analysis}
\paragraph{Comparison of ProMetaR with the generalization methods.}
\begin{table*}[!t]
     \centering
 \setlength{\tabcolsep}{8pt}
    {
    \begin{tabular}{l cc  c }
    \toprule
    \textbf{Methods}  & \textbf{Base} & \textbf{New} & \textbf{H}\\
    \midrule
    IVLP (Base)            & 82.51 & 73.36 & 77.66 \\
    Mixup            & 82.90 \textcolor{NavyBlue}{(+0.39)} & 71.45 \textcolor{Bittersweet}{(-1.91)} & 76.75 \textcolor{Bittersweet}{(-0.91)} \\
    Manifold Mixup   & 83.57 \textcolor{NavyBlue}{(+1.06)} & 73.19 \textcolor{Bittersweet}{(-0.17)} & 78.04 \textcolor{NavyBlue}{(+0.38)} \\
    EMA              & 82.30 \textcolor{Bittersweet}{(-0.21)}  & 74.15 \textcolor{NavyBlue}{(+0.79)} & 78.01  \textcolor{NavyBlue}{(+0.35)} \\
    SWA              & 83.65  \textcolor{NavyBlue}{(+1.14)} & 73.14  \textcolor{Bittersweet}{(-0.22)} & 78.04 \textcolor{NavyBlue}{(+0.38)}\\
    \rowcolor{tabhighlight}
    ProMetaR~\textbf{(Ours)}  & \textbf{84.39 \textcolor{NavyBlue}{(+1.88)}} & \textbf{76.93 \textcolor{NavyBlue}{(+3.57)}} & \textbf{80.49 \textcolor{NavyBlue}{(+2.83)}} \\
    \bottomrule
    \end{tabular}
    }
    \caption{Performance comparison of ProMetaR with the domain generalization methods on the base-to-new generalization setting. Results are averaged over 11 datasets. H refers to harmonic mean. 
    }
    \label{tab:generalization_compare}
\end{table*}
\begin{table*}[!t]
     \centering
 \setlength{\tabcolsep}{8pt}
    {
    \begin{tabular}{l cc  c }
    \toprule
    \textbf{Methods}  & \textbf{Base} & \textbf{New} & \textbf{H}\\
    \midrule
    No TaskAug             & 84.27 & 75.06 & 79.40 \\
    TaskAug: Input Mixup                  & 84.26 & 76.10 & 79.97 \\
    \rowcolor{tabhighlight}
    TaskAug: Manifold Mixup ~\textbf{(Ours)}  & \textbf{84.39} & \textbf{76.93} & \textbf{80.49} \\
    \bottomrule
    \end{tabular}
    }
    \caption{Effect of our proposed meta-regularization. Results are averaged over 11 datasets. H refers to harmonic mean. 
    }
    \label{tab:taskaug_ablation}
\end{table*}
We examine the efficacy of ProMetaR by comparing ours with data augmentation methods: Mixup~\cite{zhang2017mixup} and Manifold Mixup~\cite{verma2018manifold} and common generalization methods based on the weight averaging: exponential moving average~(EMA) and stochastic weight averaging~(SWA)~\cite{izmailov2018averaging} by applying them to the base prompt learning method, IVLP.
The results are reported in Table~\ref{tab:generalization_compare}.
Mixup slightly improves the performance on base classes with an accuracy gain of 0.39\%, but it shows the performance degradation on the new classes.
Similarly, Manifold Mixup decreases the performance on new classes with the performance gain on base classes.
These results indicate that conventional data augmentation helps improve the performance on base classes~(\textit{traditional} generalization), but it still suffers from the task overfitting problem in existing prompt learning methods to generalize on the new classes~(\textit{task} generalization).
EMA enhances new class accuracy by \textcolor{black}{+0.79\%}, at little expense of base class accuracy.
Meanwhile, SWA improves performance on base classes with an improvement of \textcolor{black}{+1.14\%}, but the average accuracy on new classes slightly decreases. 
We observe that our ProMetaR significantly outperforms both domain augmentation and generalization methods by a large margin.

\paragraph{Task augmentation.}
In Table~\ref{tab:taskaug_ablation}, we measure the performance of the model without using task augmentation~(No TaskAug), with the input Mixup~\cite{zhang2017mixup} for a task augmentation and our ProMetar that uses Manifold Mixup for the task augmentation. 
Compared to No TaskAug, task augmentation improves the performance on new classes without the loss of the performance on the base classes.
This demonstrates that using task augmentation alleviates the meta-overfitting issue by generating various virtual augmented tasks.
In addition, the task augmentation with the manifold mixup shows better performance than the input mixup with a performance gain of 0.83\% on new classes.


\end{document}